\documentclass[10pt,journal,compsoc]{IEEEtran}

\newif\ifpeerreview

\peerreviewfalse
\usepackage{subcaption}
\usepackage[labelformat=empty]{subcaption}
\usepackage[nocompress]{cite}
\usepackage{url}
\usepackage{amsmath,amssymb,graphicx}
\usepackage{booktabs}
\usepackage{xcolor}
\usepackage{multirow}
\usepackage{float}
\usepackage[switch]{lineno}
\usepackage[hidelinks]{hyperref}
\hypersetup{
    colorlinks=true, 
    urlcolor=magenta,   
}

\title{Reconstructing Satellites in 3D from Amateur Telescope Images}

\author{Zhiming Chang, Boyang Liu, Yifei Xia, Youming Guo, Boxin Shi, and He Sun% <-this % stops a space
\IEEEcompsocitemizethanks{
\IEEEcompsocthanksitem Zhiming Chang and He Sun are with the College of Future Technology and the National Biomedical Imaging Center, Peking University. Zhiming Chang is also with the College of Engineering, Peking University, Beijing, China (email: changzhiming@stu.pku.edu.cn, hesun@pku.edu.cn)
\IEEEcompsocthanksitem Yifei Xia, Boxin Shi are with the State Key Laboratory of Multimedia Information
Processing and National Engineering Research Center of Visual Technology, School of Computer Science, Peking University, Beijing, China (email: \{yfxia, shiboxin\}@pku.edu.cn)
\IEEEcompsocthanksitem Boyang Liu is with the Beijing Jingdun Technology Co., Ltd. (email: pkulby@foxmail.com)
\IEEEcompsocthanksitem Youming Guo is with the National Laboratory on Adaptive Optics and the Institute of Optics and Electronics, Chinese Academy of Sciences, Chengdu, China (email: guoyouming@ioe.ac.cn) 
\IEEEcompsocthanksitem Corresponding author: He Sun
}
}

\begin{document}

\IEEEtitleabstractindextext{%
\begin{abstract}
Monitoring space objects is crucial for space situational awareness, yet reconstructing 3D satellite models from ground-based telescope images is super challenging due to atmospheric turbulence, long observation distances, limited viewpoints, and low signal-to-noise ratios. In this paper, we propose a novel computational imaging framework that overcomes these obstacles by integrating a hybrid image pre-processing pipeline with a joint pose estimation and 3D reconstruction module based on controlled Gaussian Splatting (GS) and Branch-and-Bound (BnB) search. We validate our approach on both synthetic satellite datasets and on-sky observations of China's Tiangong Space Station and the International Space Station, achieving robust 3D reconstructions of low-Earth orbit satellites from ground-based data. Quantitative evaluations using SSIM, PSNR, LPIPS, and Chamfer Distance demonstrate that our method outperforms state-of-the-art NeRF-based approaches, and ablation studies confirm the critical role of each component. Our framework enables high-fidelity 3D satellite monitoring from Earth, offering a cost-effective alternative for space situational awareness.  Project page: \href{https://ai4scientificimaging.org/3DSatellites}{https://ai4scientificimaging.org/3DSatellites}.
\end{abstract}

}

% % Make Title
% \ifpeerreview
% \linenumbers \linenumbersep 15pt\relax 
% \author{Paper ID \paperID\IEEEcompsocitemizethanks{\IEEEcompsocthanksitem This paper is under review for ICCP 2025 and the PAMI special issue on computational photography. Do not distribute.}}
% \markboth{Anonymous ICCP 2025 submission ID \paperID}%
% {}
% \fi
\maketitle

% The first section title should be wrapped inside a \IEEEraisesectionheading as follows.
\IEEEraisesectionheading{
  \section{Introduction}\label{sec:introduction}
}
% The very first letter of the paper is a 2 line initial drop letter
% followed by the rest of the first word in caps.
\IEEEPARstart{T}{he} monitoring of man-made objects in space is an essential endeavor in space domain awareness (SDA), particularly critical for ensuring the safe operation and on-orbit servicing of civilian satellites in the era of burgeoning large-scale satellite constellations, such as SpaceX's Starlink \cite{spacex_starlink}. Within the broad spectrum of satellite monitoring activities, including detection\cite{fletcher2019feature}, categorization\cite{sanad2020statistical}, and tracking\cite{hawkins1988tracking}, 3D reconstruction stands out as it uniquely provides unparalleled insights into the real-time operational status of satellites.

\begin{figure}
    \centering
    \includegraphics[width=0.5\textwidth]{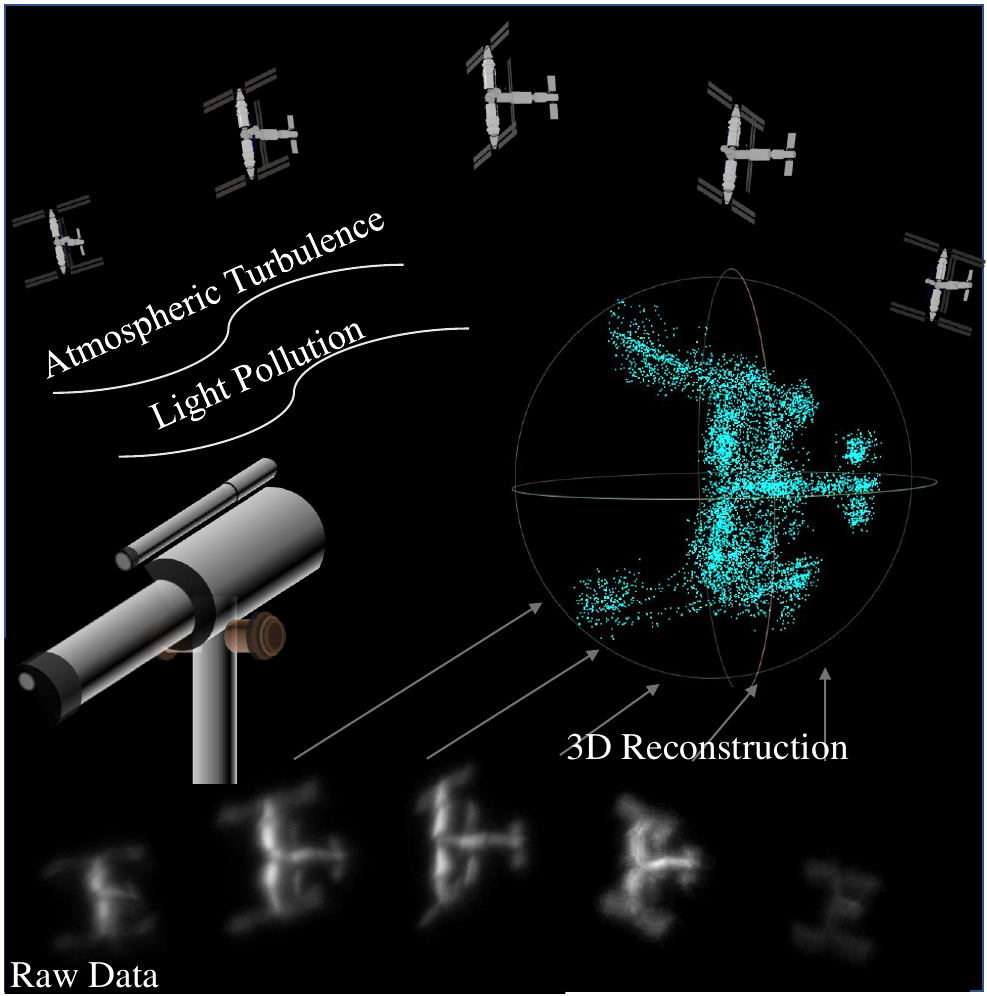}
    \caption{\textbf{Problem definition.} Our objective is to reconstruct a 3D model of a satellite in low-Earth orbit using limited, blurry, and noisy images captured by an amateur ground-based telescope, which are significantly affected by atmospheric turbulence and light pollution.}
    \label{fig:intro}
    \vspace{-15pt}
\end{figure}

To date, 3D reconstruction has only been investigated in the space-based space surveillance missions, i.e. leveraging the videos captured by a satellite to retrieve the 3D information of its neighboring space objects. 3D reconstruction from ground-based observations, despite its advantages in terms of cost-effectiveness and versatility, remains under-explored because of a series of unique challenges compared with its space-based counterparts, including but not limited to: 1) atmospheric turbulence significantly distorts images captured by ground telescopes; 2) the long observation distance complicates the accurate estimation of a satellite's pose; and 3) the rapid motion of satellites necessitates short exposure times, resulting in low signal-to-noise (SNR) ratios and constrained observational perspectives. As a result, classical 3D reconstruction approaches, no matter geometry or learning-based, fail in this challenging scenario.

In this paper, we propose an innovative computational imaging approach for reconstructing satellites in 3D using pure ground-based telescope images. Building on recent advances in imaging through turbulence and differentiable volumetric rendering, we designed a cutting-edge 3D astrophotography pipeline that integrates image restoration foundation models for pre-processing and an innovative 3D Gaussian Splatting approach for joint satellite pose and shape reconstruction. This novel pipeline for the first time enables imaging 3D space objects from limited, distorted, and noisy ground telescope observations. We validate our method using both simulated and experimental datasets. Notably, we achieve the successful reconstruction of 3D models of the International Space Station (ISS) and China's Space Station (CSS, or Tiangong Station), satellites in low Earth orbit, using solely images from an amateur 35cm-aperture telescope.

\section{Related Work}
\label{sec:related work}
\subsection{Spaceborne Satellite Monitoring}
In space missions involving surveillance and servicing, accurate 3D reconstruction and pose estimation are vital for tasks like satellite maintenance, debris removal\cite{mark2019review}, and autonomous navigation . Due to limitations in size and power for onboard instruments, these tasks often rely on monocular camera observations from companion satellites. However, traditional computer vision algorithms like SIFT and SURF\cite{lowe2004distinctive, sharma2016comparative} often struggle with the unique characteristics of space images, which typically exhibit high noise levels, simple textures, and symmetric features. 

Recent advances have leveraged deep learning to overcome these challenges. Synthetic datasets like SPEED\cite{sharma2020neural}, SPEED+\cite{park2022speed+}, and SHIRT\cite{park2023adaptive}have enabled the training of convolutional neural network (CNN)–based approaches that, when combined with predefined 3D object models (e.g., point clouds\cite{qiao2022deep} or wireframes\cite{sharma2018robust, d2014pose}), yield robust pose estimation. Furthermore, emerging research is beginning to integrate pose estimation with simultaneous 3D reconstruction to monitor unknown objects—such as space debris—from limited 2D images\cite{zhang20173d, dennison2023vision, park2024rapid}. Nevertheless, these models have primarily been tailored for the high-quality observations obtained by spaceborne telescopes. No established methods exist for reconstructing 3D satellite models purely from corrupted ground-based observations. Addressing this gap remains a critical challenge for future work in space situational awareness.

\subsection{Imaging through Turbulence}
Ground-based observations of space objects encounter significant image quality degradation due to atmospheric turbulence. Random fluctuations in the atmosphere's refractive index introduce distortions and blurring in captured images. Adaptive optics (AO) \cite{roddier1999adaptive} mitigates turbulence effects by actively correcting wavefront aberrations in real time using deformable mirrors\cite{sun2018identification, sun2020high}, but these systems are typically expensive and complex—limiting their use to large observatories.

An attractive alternative is computational correction\cite{anantrasirichai2018atmospheric,anantrasirichai2013atmospheric,mao2020image}. For example, lucky imaging\cite{law2006lucky} captures high-frame-rate video and selects brief moments when turbulence is minimal—statistically, some frames will be less affected by distortions—and then combines these "lucky" frames to produce a higher-quality image. Recent progress in turbulence simulation\cite{schwartzman2017turbulence,mao2021accelerating,chimitt2022real} has also facilitated the development of deep learning methods for turbulence mitigation. These methods train neural networks on pairs of degraded and ground-truth images to predict and correct turbulence effects
\cite{zhang2024spatio,yasarla2021learning,rai2022removing,mei2023ltt}, often surpassing classical techniques in reconstruction quality and robustness.

\subsection{3D Reconstruction and View Synthesis}
3D reconstruction is a foundational technique in computer vision and graphics that enables the modeling and visualization of complex objects. Classical approaches typically leverage geometric cues from multiple images and rely on explicit representations. Techniques such as structure-from-motion (SfM) and multi-view stereo (MVS) build explicit models\cite{schonberger2016structure}—often in the form of point clouds or meshes—by leveraging epipolar geometry and bundle adjustment\cite{fischler1981random, triggs2000bundle}. While effective in many scenarios, these techniques can struggle with texture-less surfaces, occlusions, sparse viewpoints, and complex lighting conditions, challenges often encountered in 3D satellite reconstruction.

Neural Radiance Fields (NeRF) \cite{mildenhall2021nerf, mueller2022instant} have revolutionized 3D reconstruction and view synthesis by modeling a scene's volumetric density and color using deep networks. This implicit representation enables photorealistic renderings that capture intricate lighting and transparency from sparse viewpoints\cite{fridovich2023k,cao2023hexplane}. Variants such as NeRF$--$ and BARF further incorporate motion information inspired by SfM, allowing simultaneous 3D reconstruction and pose estimation from unposed images. However, despite their ability to capture fine geometric and appearance details, NeRF-based methods typically require significant training time and offer limited ease for scene editing or post-processing.

More recent advancements have introduced 3D Gaussian splatting\cite{kerbl20233d}, a technique that represents a scene as a collection of anisotropic Gaussians rather than an implicit function. This approach yields an efficient, interpretable representation that supports real-time volumetric rendering. By "splatting" each Gaussian onto the image plane and blending overlapping contributions, the method achieves high-fidelity reconstructions at interactive frame rates. The 3D Gaussian splatting family not only delivers competitive reconstruction quality but also significantly reduces rendering time, making it particularly promising for applications that demand both accuracy and speed, as well as ease of editing\cite{chen2023gaussianeditor, fang2023gaussianeditor, huang2023point}. Multiple pose-free 3D Gaussian splatting reconstruction methods, such as SelfSplat\cite{kang2024selfsplat}, NoPoSplat\cite{ye2024no}, and FreeSplat\cite{wang2024freesplat}, have also been introduced recently. However, these techniques typically require images with rich texture information, or sometimes rely on pre-trained depth estimators (effective only for specific scenes) to provide additional regularization.

\subsection{Equatorial Mount and Meridian Flip}
\begin{figure}
    \centering
    \includegraphics[width=0.48\textwidth]{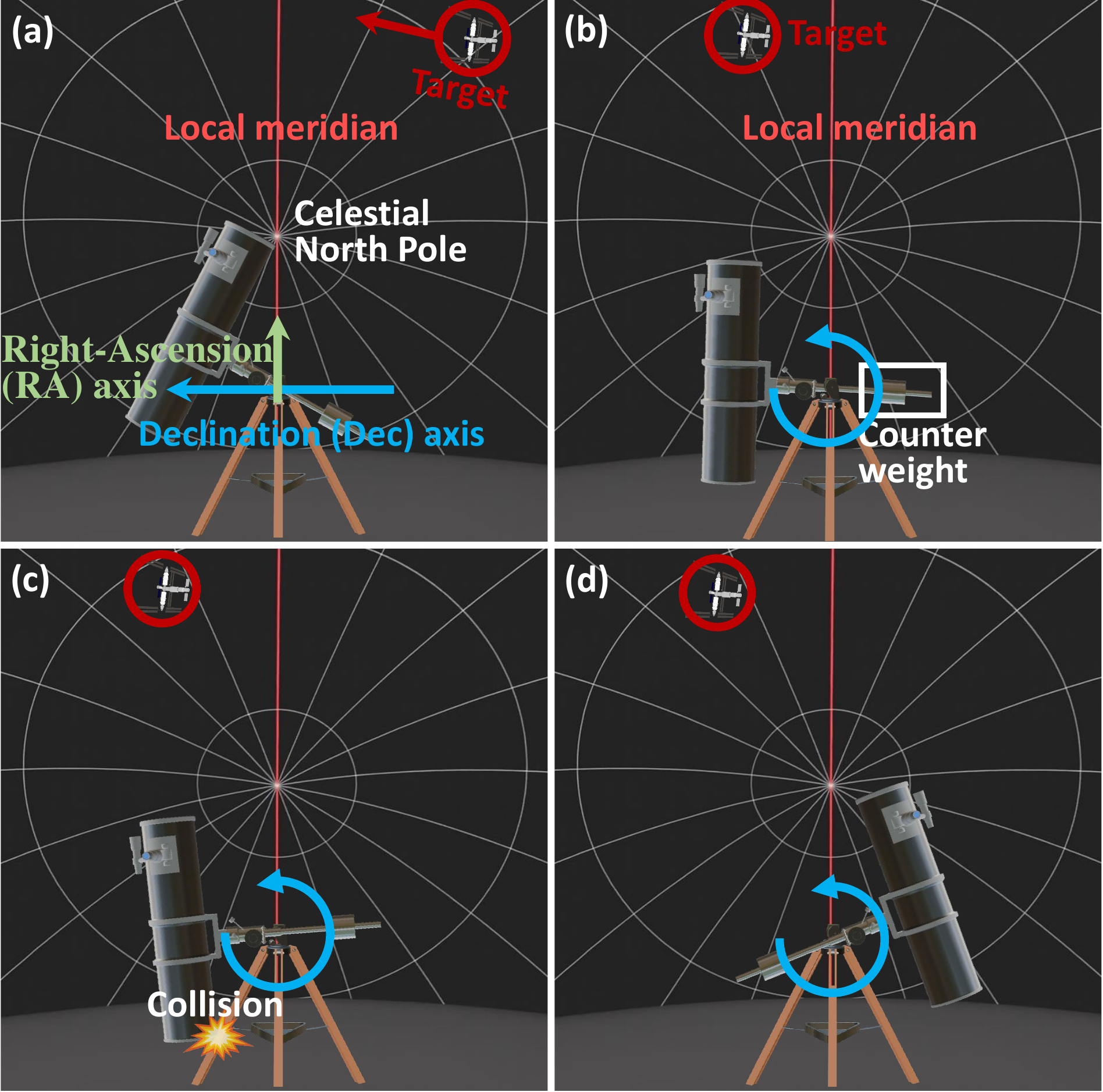}
    \caption{\textbf{Illustration of equatorial mount tracking and meridian flip.} (a) The telescope tracks a target across the celestial sphere. (b) The target reaches the local meridian. (c) Without adjustment, the telescope risks mechanical collision with the mount. (d) A meridian flip is performed, rotating the RA axis by 180$^{\circ}$ to safely continue tracking.}
    \label{fig:mount}
\end{figure}

In astronomical imaging, telescopes are commonly mounted on equatorial mounts \cite{wiki_ecs} to compensate for the Earth's rotation and the apparent motion of astronomical objects. These mounts have two orthogonal axes to enable smooth tracking: the right ascension (RA) axis, aligned with Earth's rotational axis and pointing toward the celestial north pole, and the declination (Dec) axis, which adjusts the telescope's elevation (see Fig.~\ref{fig:mount}(a)). During tracking, the RA axis rotates at either the sidereal rate or the satellite’s orbital rate to keep the target centered, while the Dec axis typically requires only minor adjustments.

However, as the target crosses the local meridian—a vertical plane passing through the zenith—the RA axis must continue rotating (Fig.\ref{fig:mount}(b)). This motion can bring the telescope dangerously close to the mount or tripod, risking mechanical collision (Fig.\ref{fig:mount}(c)). To prevent this, most equatorial mounts execute a meridian flip \cite{mcconahay2019meridian, wolfcreek_meridian_flip}, rotating the RA axis by 180$^{\circ}$ to reposition the telescope on the opposite side (Fig.~\ref{fig:mount}(d)). This operation introduces a brief interruption and a mirrored camera orientation, typically causing a pause of around 30 seconds.

While such delays are negligible for slowly drifting targets like stars or nebulae, they become problematic for fast-moving satellites, which remain visible for only 2–3 minutes per pass. Compounding this issue, the meridian region often offers the best imaging conditions due to minimal atmospheric distortion. To address these challenges in ground-based satellite tracking, we should reconfigure the mount alignment to avoid meridian flips—an approach we detail in Sec.~\ref{subsubsec:Observationscssandiss}, referred to as the "Fake Polar Axis" method.

% In astronomical imaging, telescopes are typically mounted on equatorial mounts \cite{wiki_ecs} to compensate for Earth’s rotation and ensure smooth tracking. These mounts feature two orthogonal axes: the Right Ascension (RA) axis, aligned with Earth’s rotational axis and pointing toward the celestial north pole, and the Declination (Dec) axis, which adjusts elevation (see Fig.~\ref{fig:mount}(a)). During tracking, the RA axis rotates at the sidereal rate to keep the target centered, while the Dec axis requires only minor adjustments.

% As the target crosses the local meridian—a vertical plane through the zenith—the RA axis must continue rotating (Fig.~\ref{fig:mount}(b)), which can bring the telescope dangerously close to the mount or tripod, risking collision (Fig.~\ref{fig:mount}(c)). To prevent this, most equatorial mounts perform a meridian flip\cite{mcconahay2019meridian, wolfcreek_meridian_flip}, rotating the RA axis 180° to reposition the telescope on the opposite side (Fig.~\ref{fig:mount}(d)). This introduces a brief pause and a mirrored camera orientation, typically taking about 30 seconds.

% While negligible for slow-moving targets like stars or nebulae, such interruptions are problematic for fast-moving satellites, which are visible for only 2–3 minutes per pass. Moreover, the meridian region usually offers the best imaging conditions. To overcome this, we adopt a \textit{Fake Polar Method}, eliminating meridian flips by reconfiguring mount alignment, as detailed in Sec.\ref{subsubsec:Observationscssandiss}.

\begin{figure*}
    \centering
    \includegraphics[width=0.9\textwidth]{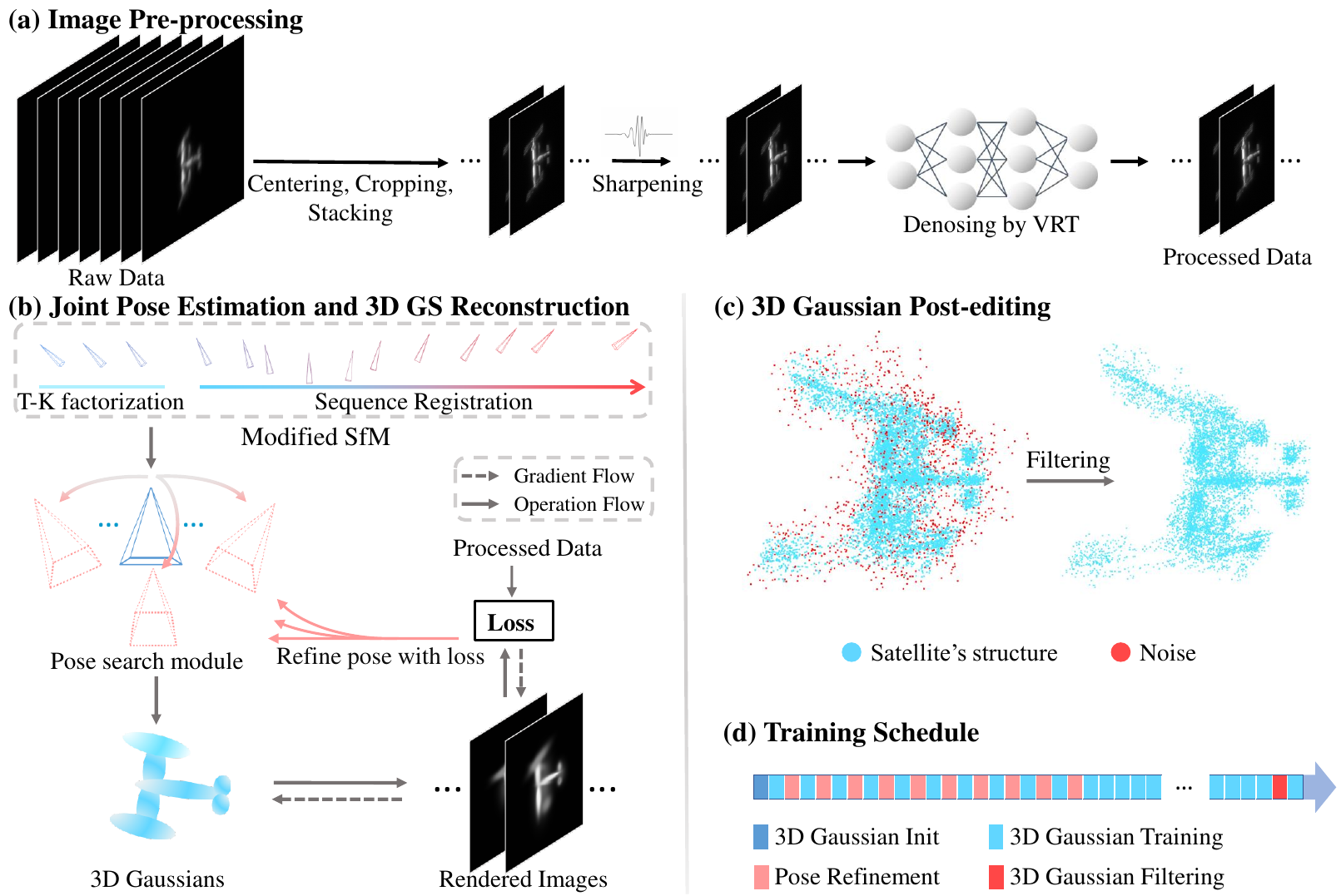}
    \caption{\textbf{Overview of the 3D satellite reconstruction method.} (a) \textbf{Image pre-processing:} A novel pipeline mitigates atmospheric turbulence and noise by combining classical signal processing with a Video Restoration Transformer (VRT). Stacking (lucky imaging) boosts the video's SNR, wavelet sharpening enhances high-frequency features, and VRT removes non-Gaussian noise and complex artifacts, significantly improving raw telescope images. (b) \textbf{Joint pose Estimation and 3D reconstruction:} Our method begins with a modified SfM for coarse pose initialization. In this approach, we employ an orthographic projection model—appropriate given the minimal perspective effects in long-distance observations—and sequentially register adjacent frames to regularize motion. Using the resulting sparse points, we initialize a 3D Gaussian point cloud and iteratively refine the Gaussian parameters and the camera poses. Pose estimation is optimized through branch-and-bound (BnB) search to minimize projection errors. Notably, we regulate the growth rate of the Gaussians to ensure they first capture the satellite's overall geometry (low-frequency information) before learning finer details (high-frequency information). This strategy is particularly effective when initial pose estimates are inaccurate. After several iterations, we halt pose updates while continuing to train the Gaussians. (c) \textbf{Post-processing:} Toward the end of training, KNN filtering removes isolated or noisy points from the Gaussian point cloud. (d) \textbf{Training schedule:} An overview of the complete training process is presented.
    }
    \label{fig:pipeline}
\end{figure*}

\section{Method}
Our method addresses the challenging problem of 3D reconstruction of space objects from unposed, monocular images captured by ground-based telescopes. In contrast to conventional 3D reconstruction tasks, our approach must contend with several unique challenges: atmospheric turbulence, extremely long observation distance, limited viewpoints, and sparse, low-quality features. As shown in Fig.~\ref{fig:pipeline}, our method overcomes these issues through two key innovations:1) a hybrid image pre-processing pipeline combining classical and deep learning techniques, and 2) a joint pose-refinement and 3D reconstruction framework using Gaussian Splatting (GS) with controlled, iterative optimization.

\subsection{Hybrid Telescope Image Pre-processing Pipeline} \label{subsec:preprocess}

Ground-based telescope images are often compromised by atmospheric turbulence, noise, and light pollution, all of which distort and blur critical features. To address these issues, we synergize classical astronomical techniques with modern foundation models in a three-stage pipeline, as detailed in Fig.~\ref{fig:pipeline} (a).

% \subsubsection{Turbulence Mitigation via Lucky Imaging}
% We begin by applying lucky imaging~\cite{autostakkert} to high-frame-rate, short-exposure image sequences, statistically selecting the top 12\% of frames based on sharpness and contrast metrics. These selected frames then undergo centroid-based alignment to correct for bulk satellite motion, followed by intensity-normalized cross-correlation to address residual sub-pixel jitter. Finally, the well-aligned frames are stacked to improve the signal-to-noise ratio.

\subsubsection{Turbulence Mitigation via Lucky Imaging}
We begin by applying lucky imaging~\cite{autostakkert} to high-frame-rate, short-exposure image sequences, statistically selecting the top 12\% of frames based on sharpness and contrast metrics. These selected frames then undergo centroid-based alignment to correct for bulk satellite motion. Each frame is subsequently divided into multiple patches, and intensity-normalized cross-correlation is applied per patch to correct residual sub-pixel jitter induced by local turbulence.
The aligned frames $\{ \widetilde{I}_t \}_{t \in \mathcal{S}}$ are stacked to suppress noise and geometric distortions. The stacked image is computed as:
\[
I_{\text{stacked}}(x,y) = \frac{1}{|\mathcal{S}|} \sum_{t \in \mathcal{S}} \widetilde{I}_t(x,y),
\]
where $\mathcal{S}$ denotes the subset of selected sharp frames. This averaging reduces zero-mean random noise by a factor of $1/\sqrt{|\mathcal{S}|}$ and improves the signal-to-noise ratio (SNR).
% Under time-varying point spread functions (PSFs) $\mathcal{P}_t$ caused by atmospheric turbulence, stacking also approximates a sharper effective PSF:
% \[
% \bar{\mathcal{P}} \approx \frac{1}{|\mathcal{S}|} \sum_{t \in \mathcal{S}} \mathcal{P}_t,
% \]
% which suppresses asymmetric aberrations and improves geometric fidelity in the final composite image. These steps are automatically performed by the widely-used astronomical processing tool AutoStakkert.\cite{autostakkert}

% \subsubsection{Multi-Scale Sharpening with Wavelet Decomposition}
% While stacking reduces noise, it may also blur high-frequency details critical for reconstruction.  To counteract these effects, we apply Gaussian wavelet decomposition to separate the image into different frequency bands, selectively amplifying high-frequency details (e.g., satellite edges and surface textures) while suppressing low-frequency background light. This effectively improve the clarity and contrast of the ground telescope images, making the features of space objects more distinguishable.
\subsubsection{Multi-Scale Sharpening with Wavelet Decomposition}
While stacking reduces noise, it may also blur high-frequency details and suppress low-contrast structures due to short exposure. To restore such details, we apply multi-scale sharpening based on Gaussian wavelet decomposition, which separates the image into different frequency bands and selectively enhances the high-frequency components.
Let \( W_j(I) \) denote the wavelet decomposition of image $I$ at scale $j$ (\( j = 1,\dots,6 \), from fine, $256 \times 256$, to coarse, $8 \times 8$). We apply a multiplicative gain factor \( \alpha_j \) to each level:
\[
\widetilde{W}_j(I) = \alpha_j \cdot W_j(I),
\]
with
\[
(\alpha_1, \alpha_2, \alpha_3, \alpha_4, \alpha_5, \alpha_6) = (1.1, 1.1, 1.1, 1.1, 1.1, 5.0).
\]
The substantially larger gain at level 6 compensates for global contrast loss and enhances the visibility of extended, low-contrast structures (e.g., solar wings).

\subsubsection{Artifact Suppression via Video Restoration Transformer}
Classical sharpening techniques sometimes introduce artifacts or over-enhance noise. To mitigate these issues, we employ the Video Restoration Transformer (VRT)~\cite{liang2022vrt}, a foundation model pre-trained on large-scale natural images, to further refine the pre-processed images. Leveraging self-attention mechanisms and temporal consistency, VRT denoises the pre-processed images, preserving essential structural details for accurate 3D reconstruction.

\subsection{Joint Pose Estimation and 3D Gaussian Splatting Reconstruction} \label{subsec:jointGS}
Once high-quality images are obtained through pre-processing, we tackle the dual problem of reconstructing the object's 3D shape and estimating its pose. This step is particularly challenging given the long observation distance, limited viewpoint, and scarce features present in our satellite images. Our solution comprises three important stages:

\subsubsection{Initialization Using Modified SfM}
\label{subsubsec:pose_init}
We begin by establishing an initial estimate of the object's pose and rough 3D structure using a modified SfM approach. In astrophotography, the satellite moves rapidly in orbit while the telescope remains fixed on Earth. To better leverage existing computer vision tools, we reframe the problem by treating the satellite as stationary and the telescope as if it "moves" around it.

Our approach first leverages pre-trained deep learning models—SuperPoint for feature extraction and SuperGlue for feature matching—to detect and pair keypoints robustly despite distortions, blur, and sparse textures inherent in telescope images. To mitigate the challenges posed by the long observation distance, we adopt an orthographic projection model\cite{julia2019orthographic} (instead of a perspective model) in the initial step, which is more robust when perspective effects are minimal. Applying this model to the first three frames yields an initial point cloud along with corresponding camera poses.

% Next, we sequentially incorporate additional frames into our model, 

Next, we revert to the full perspective camera model and sequentially integrate additional frames, as shown in the "Modified SfM" module in Fig.~\ref{fig:pipeline} (b). For each new image, we first apply the Random Sample Consensus (RANSAC) algorithm to filter out incorrect feature correspondences between adjacent images, ensuring more accurate feature matches and better associations with the 3D point cloud. Based on these refined correspondences, we then estimate the initial camera pose using the Perspective-n-Point (PnP) algorithm. Additionally, to prevent error accumulation, we perform a global bundle adjustment after each new frame is added, refining all camera poses simultaneously. Additionally, every five frames we perform triangulation and integrate new features (e.g., from the 5th and 6th frames, then the 10th and 11th, etc.) into the point cloud. This iterative process continues until all images have been processed, resulting in a complete initial 3D model and the satellite's pose estimates.

\subsubsection{Controlled Gaussian Splatting with BnB Pose Refinement}
\label{subsubsec:bnb}
The core of our reconstruction pipeline is a novel integration of 3D Gaussian splatting with iterative pose refinement~\cite{zhong2021cryodrgn2}. We begin by initializing Gaussian points from the SfM-derived point cloud, then iteratively optimize the GS parameters while holding camera poses fixed, and vice versa. This alternating optimization enables simultaneous recovery of the satellite's 3D shape and accurate camera poses despite challenges such as long observation distance, limited viewpoints, and sparse features.

During the early GS training phase, we fix the number of Gaussians and optimize only their parameters—position, covariance, opacity, and spherical harmonic coefficients (up to third order). By limiting the number of Gaussians, we force each one to cover a larger region, effectively capturing the satellite’s coarse geometry while preventing overfitting to high-frequency artifacts when pose estimates are imprecise. After a set number of epochs, we automatically clone and split the Gaussians to increase their count, then resume training on the expanded set with the count held fixed. This controlled growth strategy enables the model to first learn coarse geometry and gradually capture finer details.

In the first 10,000 epochs of training, we refine the camera poses through a Branch-and-Bound (BnB) search performed every 1,000 epochs. Candidate poses are generated by applying controlled perturbations to the current estimate. Specifically, for rotation, we sample a random rotation angle, \( \Delta \theta \), and a random axis, \( \vec{n} \), then compute the rotation perturbation using Rodrigues' formula:
\begin{equation}
R_{\text{noise}} = I + \sin(\Delta \theta) [\vec{n}]_{\times} + (1 - \cos(\Delta \theta)) [\vec{n}]_{\times}^2,
 \label{eq:rotation_noise}
 \end{equation}
where \( [\vec{n}]_{\times} \) is the skew-symmetric matrix of \( \vec{n} \). Translation perturbations are drawn from a zero-mean Gaussian distribution:
\begin{equation}
T_{\text{noise}} \sim \mathcal{N}(0, \sigma_T^2),
\label{eq:translation_noise}
\end{equation}
Each candidate pose is then given by:
\begin{equation}
R_i = R \cdot R_{\text{noise}}, \quad T_i = T + T_{\text{noise}},
\label{eq:candidate_generation}
\end{equation}
At each BnB stage, we sample 200 pose candidates (example camera poses are shown in red in Fig.~\ \ref{fig:pipeline}(b)). We evaluate these candidates using a photometric consistency loss and select the pose \( (R_{\text{best}}, T_{\text{best}}) \) that minimizes this loss. After each stage, we shrink the translation range to one-fourth and the rotation range to one-half of the previous interval, centering both on the current camera pose estimate. This refinement schedule ensures stable convergence within ten rounds. The best pose is then passed to the next GS training phase. By alternating between Gaussian splatting with adaptive point growth and BnB-based pose refinement, our method jointly converges to accurate 3D structure and camera poses.

Our implementation builds on the original 3D Gaussian Splatting framework~\cite{kerbl20233d}, using the same hyper-parameter set across all experiments—synthetic data, CSS, and ISS—without scene-specific tuning, and keeping all other settings unchanged except for the modifications described above.

\subsubsection{Explicit Post-processing of the Learned Point Cloud} 
Because the reconstructed 3D Gaussians may include stray points from background noise, we further apply a statistical filtering step to improve model accuracy. As illustrated in Fig.~\ref{fig:pipeline}(c), we examine each Gaussian’s spatial location (ignoring its appearance and shape attributes), compute the mean distance to its $k$ nearest neighbors, and remove any point whose mean distance exceeds the global mean plus one standard deviation ($\mu + \sigma$). After filtering, we resume Gaussian splatting training on the refined 3D Gaussians for an additional 500 epochs to repair discontinuities and further improve reconstruction quality.

Fig.~\ref{fig:pipeline} (d) summarizes the complete training schedule of our 3D reconstruction pipeline—including initialization, GS training, pose refinement, and explicit filtering. Our integrated approach, enabled by careful algorithmic design, effectively overcomes multiple fundamental challenges of reconstructing space objects from degraded ground-based telescope images.

\setlength{\tabcolsep}{3pt}
\begin{table*}[htbp]
  \centering
  
  \caption{\textbf{Quantitative comparison of image pre-processing quality between our method and various baselines on synthetic data.} The best results in each column are in \textbf{bold}, and second-best are \underline{underlined}.Our method achieves the highest PSNR and SSIM metrics. For a detailed explanation of each method, please refer to the Fig.~\ref{fig:preprocess}'s caption.
  % Performance metrics across different image preprocessing methods. Dirty images represents the raw data. Restormer and Video Rstortation Transformer(VRT) are pretrained deep learning models for image restoration, while selfDeblur\cite{ren2020neural} is a self-supervised neural blind deconvolution method. Stacked refers to the frame alignment and averaging process from lucky imaging, and Sharpened applies wavelet-based enhancement. Best results are in \textbf{bold}, and second-best are \underline{underlined}. Our method demonstrates the best performance in both PSNR and SSIM metrics.
  }
   % Three standard imaging metrics, PSNR, SSIM and LPIPS, for both training and validation views are reported.
  \begin{tabular}{lccccccccccc}
    \toprule
    & \multicolumn{3}{c}{Simulation 1} & \multicolumn{3}{c}{Simulation 2} & \multicolumn{3}{c}{Simulation 3} \\
    \cmidrule(r){2-4} \cmidrule(r){5-7} \cmidrule(r){8-10}
    Method & \scriptsize{PSNR\textsuperscript{$\uparrow$}} & \scriptsize{SSIM\textsuperscript{$\uparrow$}} & \scriptsize{LPIPS\textsuperscript{$\downarrow$}} & \scriptsize{PSNR\textsuperscript{$\uparrow$}} & \scriptsize{SSIM\textsuperscript{$\uparrow$}} & \scriptsize{LPIPS\textsuperscript{$\downarrow$}} & \scriptsize{PSNR\textsuperscript{$\uparrow$}} & \scriptsize{SSIM\textsuperscript{$\uparrow$}} & \scriptsize{LPIPS\textsuperscript{$\downarrow$}}\\
    \midrule
    Dirty Images & 23.28 & 0.07 & 0.41 & 23.02 & 0.07 & 0.36 & 23.16 & 0.07 & 0.36  \\

    Restormer & 24.83 & 0.18 & \textbf{0.09} & 24.76 & 0.18 & \textbf{0.10} & 25.06 & 0.18 & \textbf{0.10}\\
    
    Stacking & 27.00 & 0.41 & 0.15 & 27.15 & 0.41 & 0.17 & 27.75 & 0.41 & 0.16 \\
    
    Stacking + SelfDeblur & 23.89 & 0.46 & \underline{0.10} & 24.39 & 0.45 & \textbf{0.10} & 24.39 & 0.45 & \textbf{0.10} \\

    Stacking + Restormer & 26.78 & 0.39 & 0.14  & 26.93 & 0.38 & 0.16 & 27.49 & 0.39 & 0.15\\
    
    Stacking + Sharpening & \underline{27.60} & \textbf{0.53} & \underline{0.10} & \underline{27.92} & \underline{0.53} & \underline{0.11} & \underline{28.3} & \underline{0.52} & 0.11 \\
    
    Ours (Stacking + Sharpening + VRT) & \textbf{27.74} & \textbf{0.53} & 0.11 & \textbf{28.02} & \textbf{0.55} & 0.13 & \textbf{28.47} & \textbf{0.54} & 0.11\\
    
    \bottomrule
  \end{tabular}
  \label{tab:preprocess}
\end{table*}

\begin{table}
  \centering
  \caption{
  \textbf{Quantitative comparison of pose estimation accuracy (Absolute Trajectory Error, ATE) for various methods on synthetic data.} "N/A" indicates failure to converge. The top half presents results for pose initialization, while the bottom half shows outcomes after joint 3D reconstruction and pose refinement. Notably, our pose initialization method is the only one that produces reasonable initial estimates, and our overall approach achieves the highest accuracy after joint optimization.
  % Comparison of Absolute Trajectory Error (ATE) across different pose estimate methods on Simulation Datasets. The upper part presents the results of pose initialization, where "N/A" indicates that the method failed to produce a valid result. Only our method achieves a trajectory closer to the ground truth, as illustrated in Fig.~\ref{fig:pose}. The lower part shows the pose refinement results after joint pose refinement and 3D reconstruction, where \textbf{all three methods are initialized with our trajectory}. Ultimately, our BnB-based method achieves the most accurate pose estimation, closest to the ground truth.
  }
  \begin{tabular}{lc}
    \toprule
    Method & Absolute Trajectory Error\textsuperscript{$\downarrow$} \\
    \midrule
    \textbf{Pose Initialization} \\
    Colmap / Hloc / Detector-Free SfM & N/A \\
    Dust3r  & 138.81 \\
    RayDiffusion  & 138.76 \\
    Ours & \textbf{2.61} \\
    \midrule
    \textbf{After pose Refinement} \\
    After BnB & \textbf{2.43}\\
    NeRF$--$ & 2.57 \\
    BARF (Constrain Sampling) & 2.54 \\
    \bottomrule
  \end{tabular}
  \label{tab:pose}
\end{table}

\setlength{\tabcolsep}{1pt}
\begin{table*}[htbp]
  \centering
  \caption{\textbf{Quantitative comparison of 3D reconstruction accuracy between our method and NeRF-based approaches.} Results are evaluated using three simulation experiments and two on-sky observations, China's Space Station (CSS) and the International Space Station (ISS). PSNR, SSIM, and LPIPS metrics are computed by comparing novel views with Blender renderings for simulated data and with pre-processed reserved validation views for on-sky data. Best results in each column are shown in \textbf{bold}, and second-best values are \underline{underlined}.
  % Comparison of our method and NeRF-based methods on three simulations and on-sky observations of China's Space Station (CSS) and the International Space Station (ISS), showing validation view results, compared with the results directly rendered in Blender for the simulated data, which are free of noise and turbulence, and with the on-sky data compared to the preprocessed images. Best results are in \textbf{bold}, and second-best are \underline{underlined}.
  }
   % Three standard imaging metrics, PSNR, SSIM and LPIPS, for both training and validation views are reported.
  \begin{tabular}{lcccccccccccccccccccc}
    \toprule
    & \multicolumn{4}{c}{Simulation 1} & \multicolumn{4}{c}{Simulation 2} & \multicolumn{4}{c}{Simulation 3} & \multicolumn{3}{c}{CSS} & \multicolumn{3}{c}{ISS} \\
    \cmidrule(r){2-5} \cmidrule(r){6-9} \cmidrule(r){10-13} \cmidrule(r){14-16} \cmidrule(r){17-19}
    Method & \scriptsize{PSNR\textsuperscript{$\uparrow$}} & \scriptsize{SSIM\textsuperscript{$\uparrow$}} & \scriptsize{LPIPS\textsuperscript{$\downarrow$}} & 
    \scriptsize{CD\textsuperscript{$\downarrow$}} &
    \scriptsize{PSNR\textsuperscript{$\uparrow$}} & \scriptsize{SSIM\textsuperscript{$\uparrow$}} & \scriptsize{LPIPS\textsuperscript{$\downarrow$}} & 
    \scriptsize{CD\textsuperscript{$\downarrow$}} &
    \scriptsize{PSNR\textsuperscript{$\uparrow$}} & \scriptsize{SSIM\textsuperscript{$\uparrow$}} & \scriptsize{LPIPS\textsuperscript{$\downarrow$}} & 
    \scriptsize{CD\textsuperscript{$\downarrow$}} &
    \scriptsize{PSNR\textsuperscript{$\uparrow$}} & \scriptsize{SSIM\textsuperscript{$\uparrow$}} & \scriptsize{LPIPS\textsuperscript{$\downarrow$}} & \scriptsize{PSNR\textsuperscript{$\uparrow$}} & \scriptsize{SSIM\textsuperscript{$\uparrow$}} & \scriptsize{LPIPS\textsuperscript{$\downarrow$}}\\
    \midrule
    Ours & \textbf{25.32} & \textbf{0.90} & \textbf{0.16} & \textbf{0.07} & \underline{27.53} & \textbf{0.91} & \underline{0.14} & \textbf{0.08} & \textbf{28.10} & \textbf{0.91} & \textbf{0.11} & \textbf{0.07} & \textbf{34.68} & \textbf{0.96} & \textbf{0.03} & \textbf{36.87} & \textbf{0.98} & \textbf{0.02}\\
    NeRF$--$ & 16.09 & 0.53 & 0.37 & 0.34 & 18.06 & \underline{0.88} & 0.29 & \underline{0.34} & 16.24 & \underline{0.76} & 0.36 & 0.42 & 15.09 & 0.67 & 0.34 & 29.05 & \underline{0.92} & 0.10\\
    BARF & 17.64 & \underline{0.55} & \underline{0.40} & 0.38 & 17.28 & 0.47 & 0.35 & 0.35 & \underline{22.82} & 0.53 & 0.18 & 0.43 & 23.52 & 0.77 & 0.15 & 30.85 & 0.88 & 0.12 \\
    BARF (Con.) & \underline{24.20} & 0.54 & \textbf{0.16} & \underline{0.34} & \textbf{28.67} & 0.62 & \textbf{0.11} & 0.34 & 22.07 & 0.56 & \underline{0.14} & \underline{0.37} & \underline{31.04} & \underline{0.87} & \underline{0.06} & \underline{31.53} & \underline{0.92} & \underline{0.06}\\
    % SCNeRF & 18.19 & \underline{0.78} & 0.37 & 0.65 & 18.75 & 0.79 & 0.39 & \underline{0.29} &  19.28 & 0.69 & 0.34 &\underline{0.34} & 17.62 & 0.75 & 0.27 & 18.14 & 0.76 & 0.38\\
    \bottomrule
  \end{tabular}
  \label{tab:result}
\end{table*}

\begin{figure*}
    \centering
    \includegraphics[width=\textwidth]{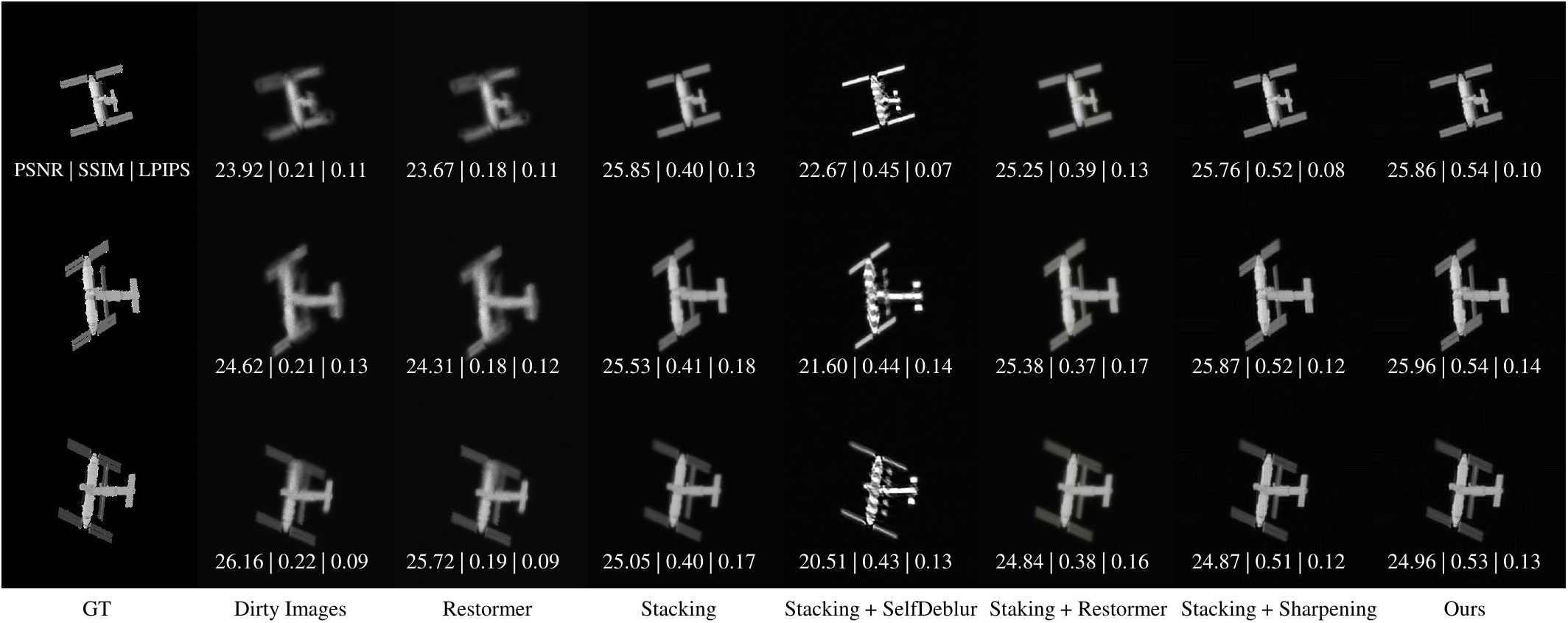}
    \caption{\textbf{Comparison of pre-processed images obtained using our method and various baselines on synthetic data.} “Dirty images” denote the raw data; “Restormer” is a pre-trained deblurring network, selfDeblur\cite{ren2020neural} is a self-supervised blind deconvolution method, and “VRT” is a pre-trained multi-task video restoration network used for denoising. “Stacking” refers to the frame alignment and averaging process from lucky imaging, while “Sharpening” applies wavelet-based enhancement. Our method yields the most faithful reconstruction.
    % Comparison of image preprocessing results using various methods. "GT" are rendered in Blender, which are free of noise and turbulence. "Dirty Images" are raw simulated captures degraded by atmospheric turbulence, noise and light pollution. Restormer, a pretrained deep model, slightly enhances sharpness but leaves noticeable blur. "Stacked" applies lucky imaging by aligning and averaging multiple frames, effectively reducing noise but still lacking detail. "Stacked-SelfDeblur" introduces self-supervised blind deconvolution to boost contrast, though at the cost of artifacts. "Stacked-Restormer" and "Stacked-Sharpened" further apply deep restoration and wavelet enhancement, improving clarity but introducing noise or over-sharpening. In contrast, our method, which integrates the Video Restoration Transformer (VRT), yields the most faithful reconstruction, achieving the best quantitative scores (PSNR, SSIM). Detailed metrics are listed in Table~\ref{tab:preprocess}.
    }
    \label{fig:preprocess}
\end{figure*}

\begin{figure}
    \centering
    \includegraphics[width=0.47\textwidth]{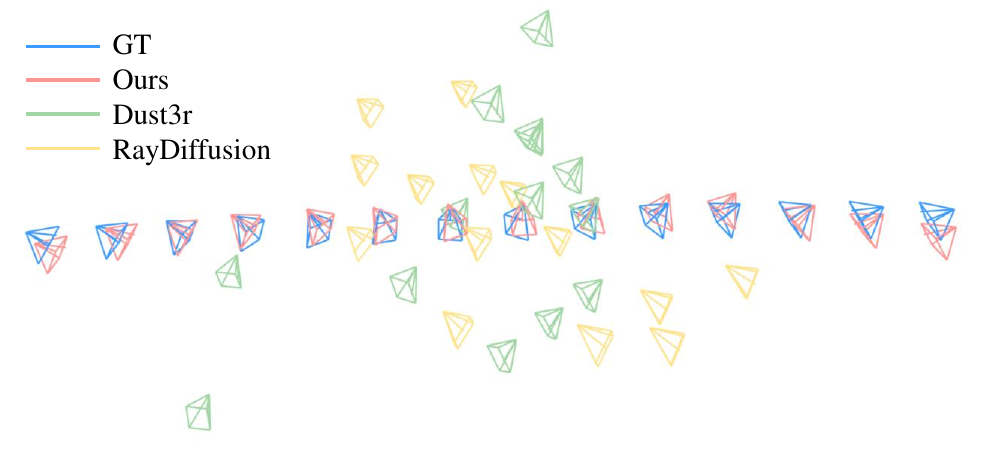}
    \caption{
    \textbf{Comparison of estimated poses generated by our method and various baselines.} Methods that fail to converge, such as Colmap, Hloc, and Detector-free SfM, are omitted. Our approach achieves the highest pose initialization accuracy.
    %, and its alignment with the ground truth further improves after joint 3D GS reconstruction and BnB-based pose refinement.
    % Comparison of estimated poses across different methods. The estimated poses from Ours, Dust3r, and RayDiffusion have been aligned with GT by compensating for differences in position, rotation, and scale. Despite this alignment, our method exhibits a pose distribution that remains significantly closer to GT, while other methods show larger deviations. This observation is further supported by Table~\ref{tab:pose}, where our approach achieves the lowest Absolute Trajectory Error (ATE), indicating superior accuracy in pose estimation.
    }
    \label{fig:pose}
\end{figure}

\begin{figure*}
    \centering
    \includegraphics[width=0.9\textwidth]{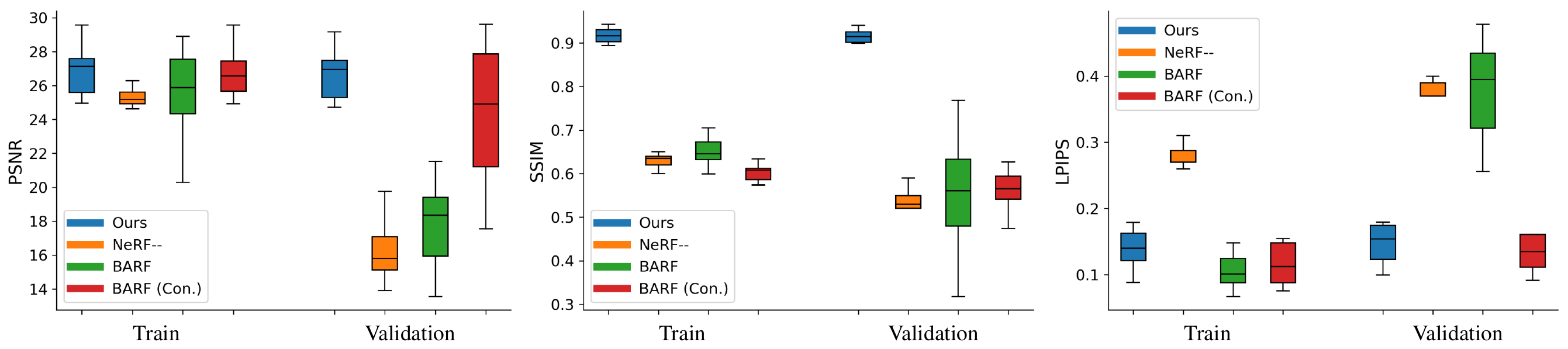}
    \caption{\textbf{Box plots comparing PSNR, SSIM, and LPIPS results for our method versus NeRF-based methods on Simulation 1.} Our approach consistently delivers stable performance across both training and validation sets. In contrast, even with the constrained sampling strategy illustrated in Fig.~\ref{fig:sample}, NeRF-based methods exhibit significant discrepancies between the training and validation sets, indicating lower robustness.
    % This box plot compares the PSNR, SSIM, and LPIPS results for our method against NeRF-based methods on Simulation 1. Our approach achieves stable results on both the training and validation sets. While NeRF-based methods exhibit a significant discrepancy between the overall training and validation sets, indicating lower robustness. It is worth noting that our method achieves impressive results on the SSIM metric, both on the training set and the test set. This success stems from our post-processing strategy, which significantly enhances the similarity of images by filtering out background noise. We have also validated this finding through ablation studies (Referring to the "No Filter" section in Table~\ref{tab:ablation}).
    }
    \label{fig:metrics}
\end{figure*}

\section{Experiments}

In this section, we evaluate our method using both simulated and real on-sky data. We first benchmark our pre-processing method against multiple image deblurring baselines, and then compare our 3D reconstruction performance with that of cutting-edge NeRF-based baselines\cite{wang2021nerf, lin2021barf}. All pose-free 3D GS baselines are excluded from our discussion, as experiments indicate that none are applicable due to the sparse textures inherent in astrophotography. We quantitatively assess reconstruction accuracy using standard imaging and 3D metrics—SSIM, PSNR, LPIPS, and Chamfer Distance (CD)—and evaluate camera pose accuracy via the Absolute Trajectory Error (ATE). Our method demonstrates superior quality and robustness on synthetic ground-based telescope images as well as on real observations of China's Space Station (CSS) and the International Space Station (ISS).

\begin{figure}
    \centering
    \includegraphics[width=0.48\textwidth]{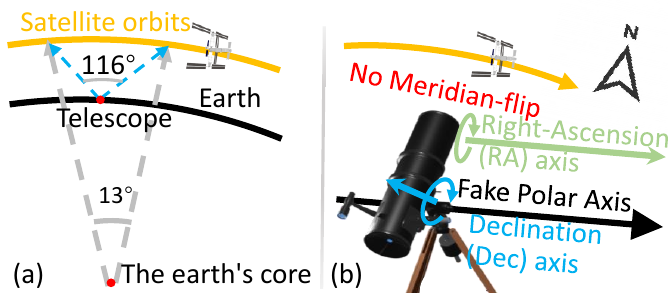}
    \caption{\textbf{Satellite tracking geometry and the Fake Polar method:} (a) A 13° ground track corresponds to a 116° apparent angular sweep from the telescope. (b) Tilting the mount’s polar axis into the orbital plane avoids meridian flip and shifts tracking motion primarily to the Dec axis.}
    \label{fig:rebuttal}
    \vspace{-12pt}
\end{figure}

\subsection{Simulation Results}
\subsubsection{Synthetic Ground Telescope Satellite Images}

% We generated three distinct 3D satellite models using Blender, each approximately 60 meters in size, and simulated ground telescope images for each. Placed in a low-Earth orbit at 650 km altitude, each satellite is visible from a stationary ground telescope in a 13$^{\circ}$ segment, \zhiming{which corresponds to an apparent 116$^{\circ}$ relative rotation between the satellite and the telescope(see Fig.~\ref{fig:rebuttal} (a)).} Our virtual telescope—with a 0.35-meter aperture and a 3.2-meter focal length—rendered 700 high-quality images per model. To simulate realistic observing conditions, each clean image was augmented with 20 different atmospheric turbulence settings (Fried parameter between 0.07 and 0.35 m) and 5–7\% additional background intensity due to light pollution. This resulted in a dataset of 14,000 distorted frames per model, each with a resolution of 512 $\times$ 512 pixels, which we will publicly release after the paper is accepted.

We generated three distinct 3D satellite models using Blender, each approximately 60 meters in size, and simulated ground telescope images for each. Placed in a low-Earth orbit at 650 km altitude, each satellite is visible from a stationary ground telescope in a 13$^{\circ}$ segment, which corresponds to an apparent 116$^{\circ}$ relative rotation between the satellite and the telescope (see Fig.~\ref{fig:rebuttal} (a)). Our virtual telescope—with a 0.35-meter aperture and a 3.2-meter focal length—rendered 700 high-quality images per model. To simulate realistic observing conditions, we introduced a full degradation model incorporating both atmospheric turbulence and light pollution. Specifically, each clean image was degraded as:
\[
I_{\text{noisy}} = (I_{\text{clean}} \ast \text{PSF}_{\text{turb}}) + B + n_{\text{ph}} + n_{\text{read}}.
\]
Here, \( \text{PSF}_{\text{turb}} \) is the optical point spread function degraded by atmospheric turbulence, simulated using the Phase-to-Space Transform (P2S) method~\cite{mao2021accelerating}. P2S approximates Kolmogorov turbulence by sampling Zernike polynomials and PCA bases learned from real turbulence data. The turbulence strength is controlled via the Fried parameter \( r_0 \), which we sampled uniformly from the range [0.07m, 0.35m], equivalent to restricting the turbulence integration height to roughly the lowest 10 km of the atmosphere. Light pollution was modeled as a combination of additive background bias \( B \sim \mathcal{U}(0.05 \, I_{\text{clean}}^{\text{max}}, 0.07 \, I_{\text{clean}}^{\text{max}}) \), signal-dependent photon noise \( n_{\text{ph}} \sim \mathcal{N}(0, \sigma_1^2 (I_{\text{turb}} + B)^2) \), and signal-independent readout noise \( n_{\text{read}} \sim \mathcal{N}(0, \sigma_2^2) \), where \( I_{\text{turb}} = I_{\text{clean}} \ast \text{PSF}_{\text{turb}} \). This procedure resulted in a dataset of 14,000 degraded frames per model, each at a resolution of 512 $\times$ 512 pixels. All synthetic data and code are available on our project page (see Abstract).

\subsubsection{Raw Telescope Image Pre-processing}
We first validate our pre-processing pipeline on the simulated data. Following our pipeline—lucky imaging and stacking (every 100 images), wavelet-based multi-scale sharpening, and VRT denoising—we reduce the 14,000 distorted frames to 140 high-quality images per satellite model. We compared our method against other cutting-edge deblurring algorithms. Table~\ref{tab:preprocess} summarizes the evaluation metrics (SSIM, PSNR, LPIPS) across various preprocessing methods. The results indicate that our proposed pipeline (Stacking + Sharpening + VRT) consistently surpasses pure deep learning approaches (e.g., Restormer\cite{zamir2022restormer}), traditional techniques (e.g., Stacking + Sharpening), self-supervised methods, and other hybrid strategies, specifically in PSNR and SSIM metrics.

\begin{figure}
    \centering
    \includegraphics[width=0.3\textwidth]{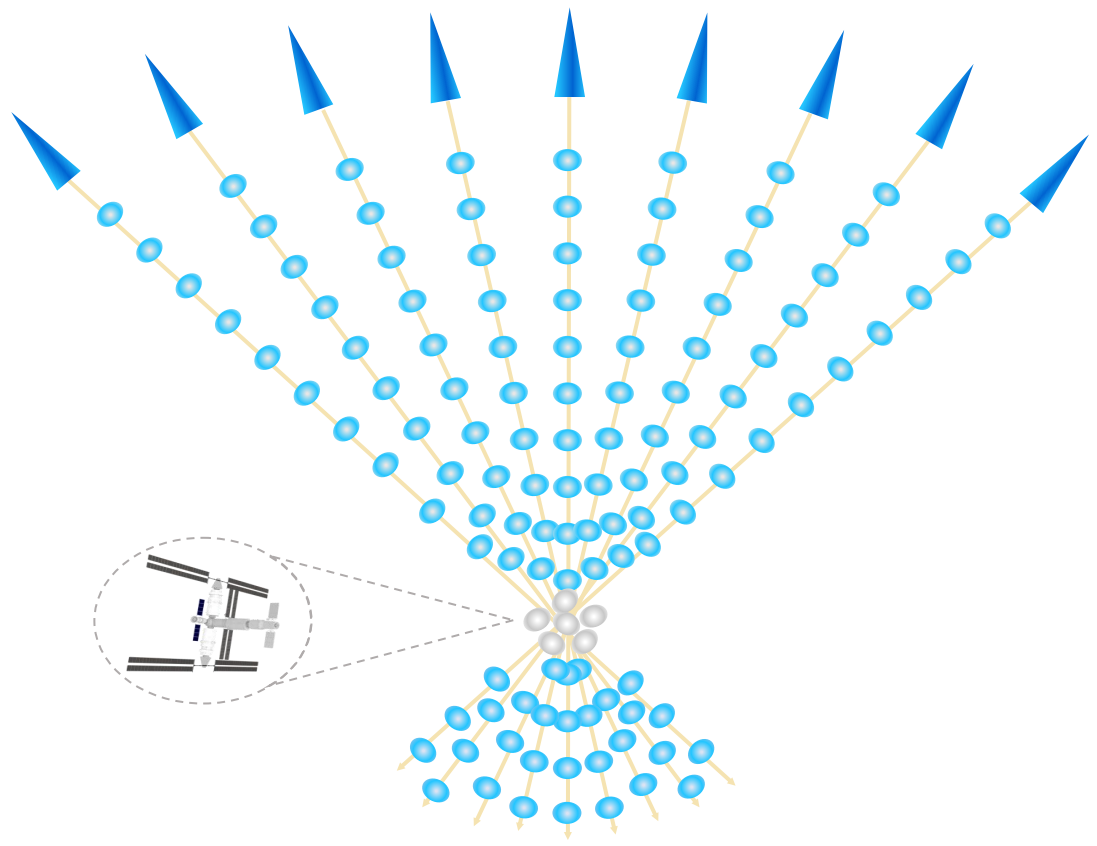}
    \caption{\textbf{Constrained sampling strategy for BARF.} Dark blue triangles represent cameras, yellow rays denote light paths, light blue dots indicate the original uniformly sampling region, and gray dots mark the constrained sampling region suggested by the initial camera-to-target distance estimate.
    % Custom NeRF sampling strategy. Deep blue triangles indicate cameras, yellow rays are light rays, light blue sphere show uniform sampling points, and metallic sphere reflect our tailored sampling based on camera distance to the point cloud center.
    }
    \label{fig:sample}
\end{figure}

\begin{figure*}
    \centering
    % \vspace{2mm}
    {\raggedright (a) 3D Point Cloud Reconstruction Results \par}
    % \vspace{1mm}

    \noindent
    \begin{minipage}{0.49\textwidth}
    \centering
        Simulation 1
        \end{minipage}%
    \hfill
    \begin{minipage}{0.49\textwidth}
        \centering
        Simulation 2
    \end{minipage}
    \begin{minipage}{\textwidth}
        \begin{minipage}{0.5\textwidth}
            \begin{minipage}{0.5\textwidth}
                \includegraphics[width=0.5\linewidth]{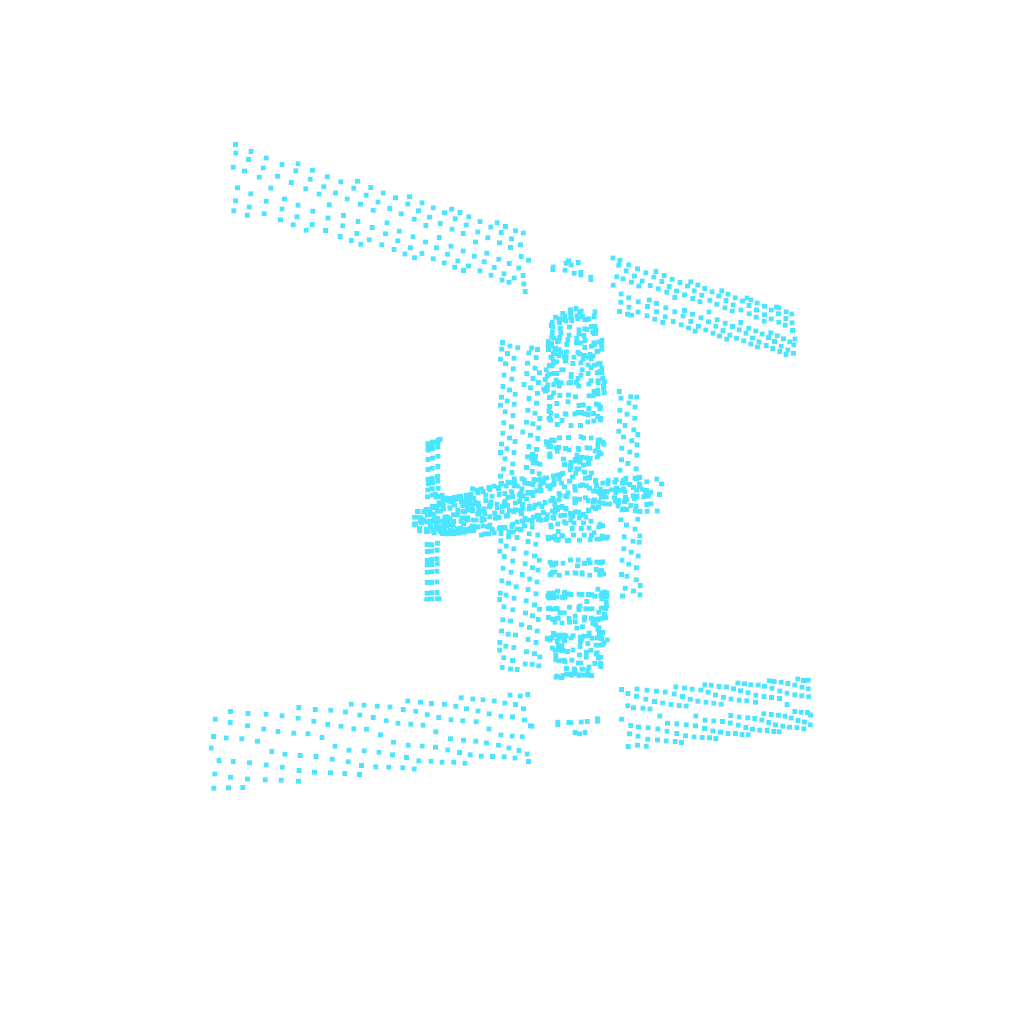}%
                \includegraphics[width=0.5\linewidth]{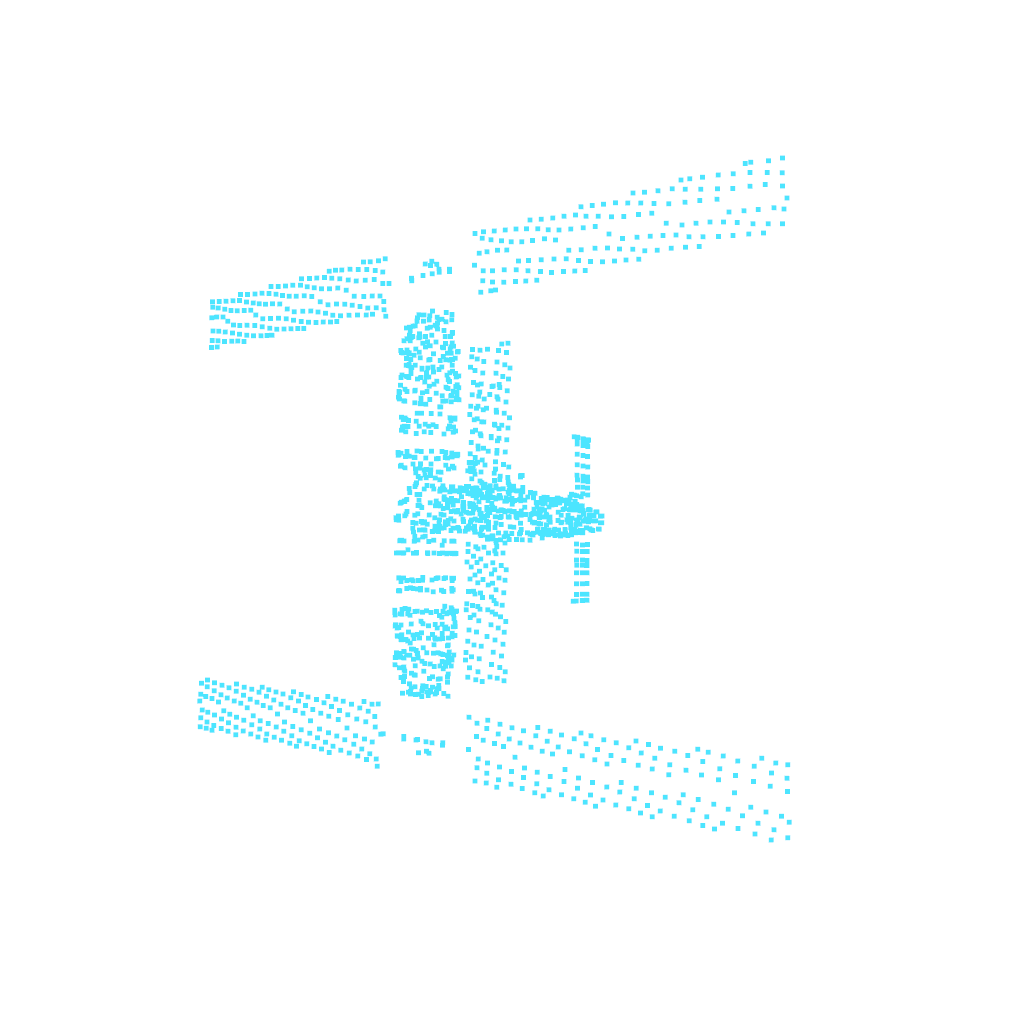}
                \subcaption*{GT}
            \end{minipage}%
            \begin{minipage}{0.5\textwidth}
                \includegraphics[width=0.5\linewidth]{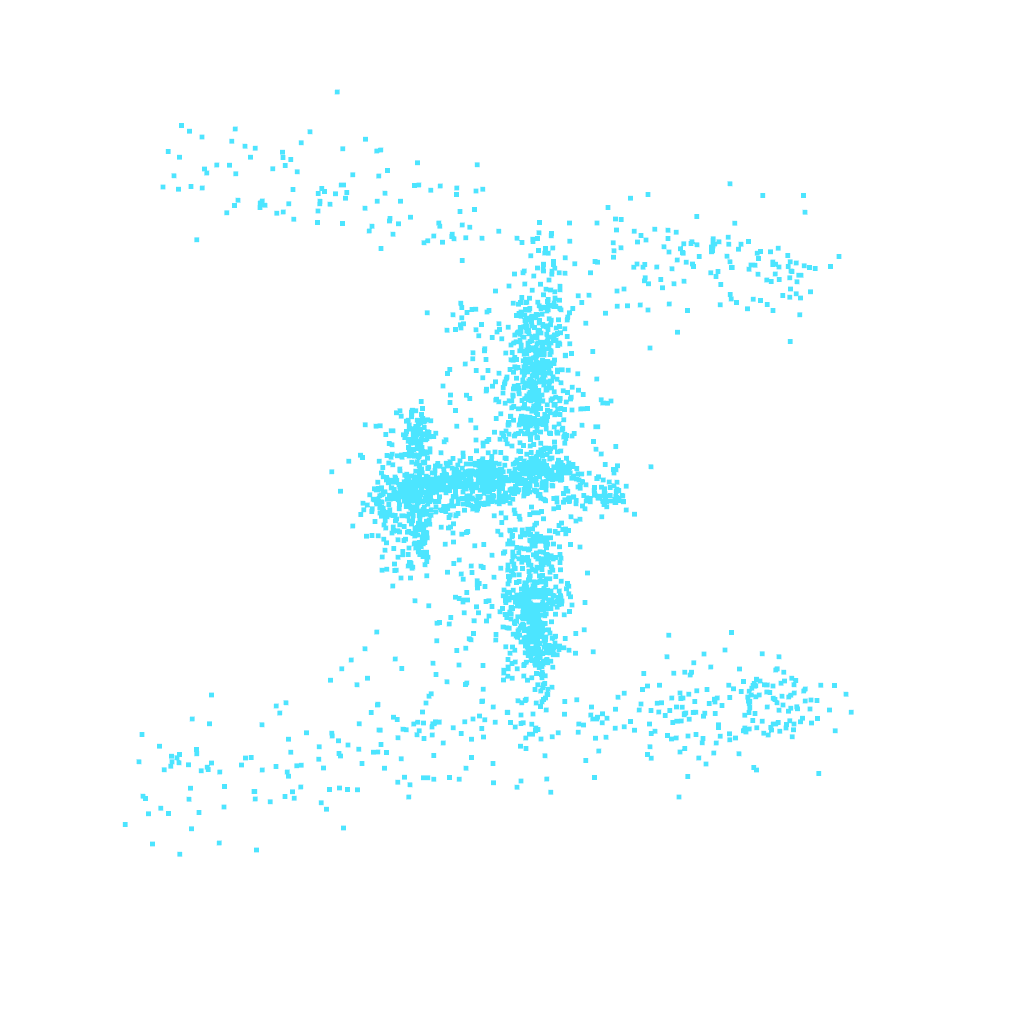}%
                \includegraphics[width=0.5\linewidth]{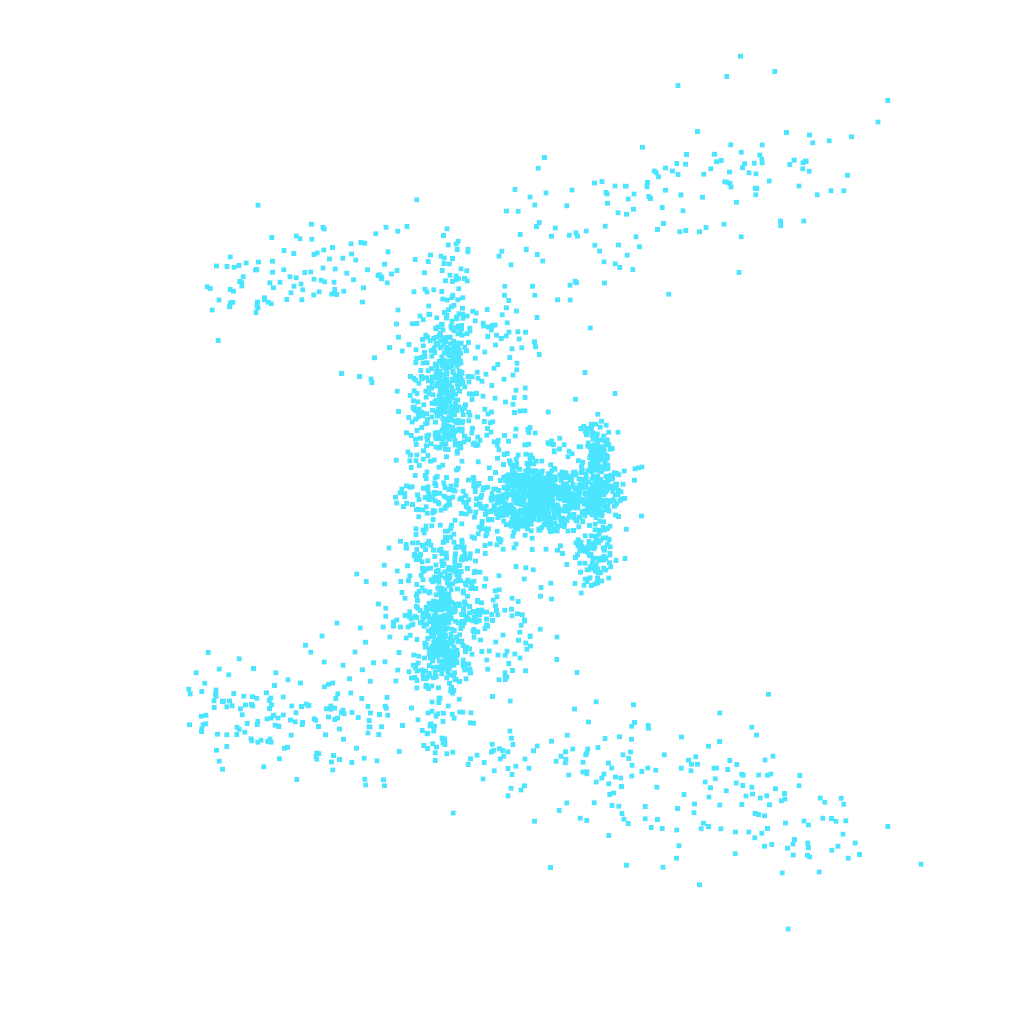}
                \subcaption*{Ours}
            \end{minipage}%
        \end{minipage}%
        \begin{minipage}{0.5\textwidth}
            \begin{minipage}{0.5\textwidth}
                \includegraphics[width=0.5\linewidth]{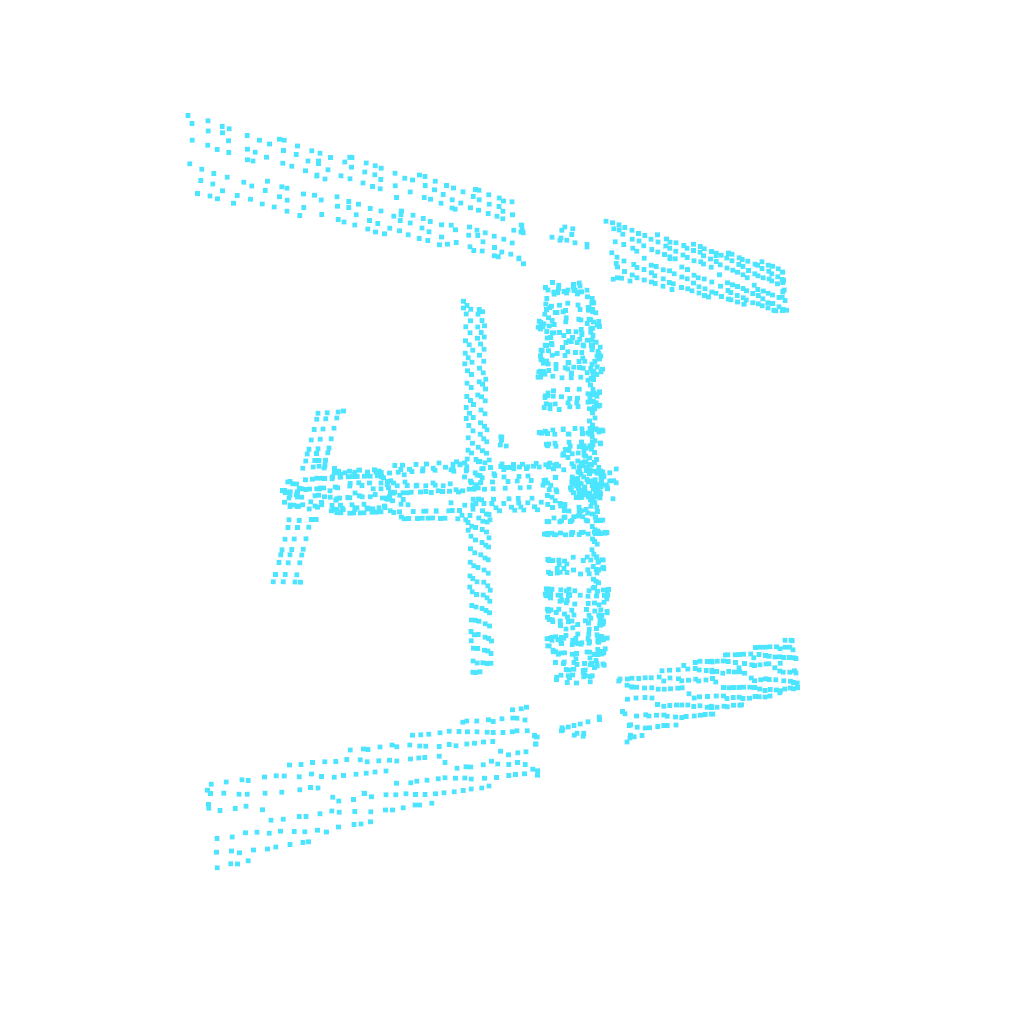}%
                \includegraphics[width=0.5\linewidth]{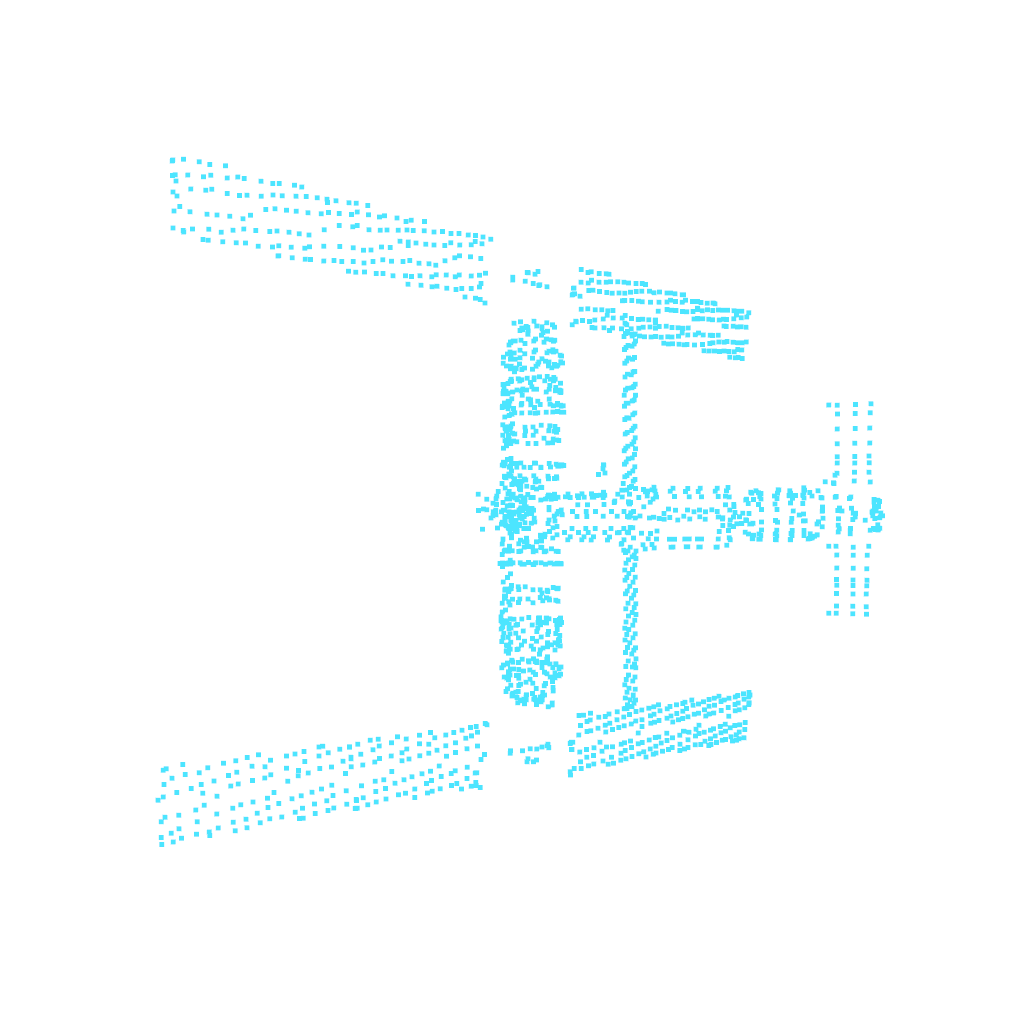}
                \subcaption*{GT}
            \end{minipage}%
            \begin{minipage}{0.5\textwidth}
                \includegraphics[width=0.5\linewidth]{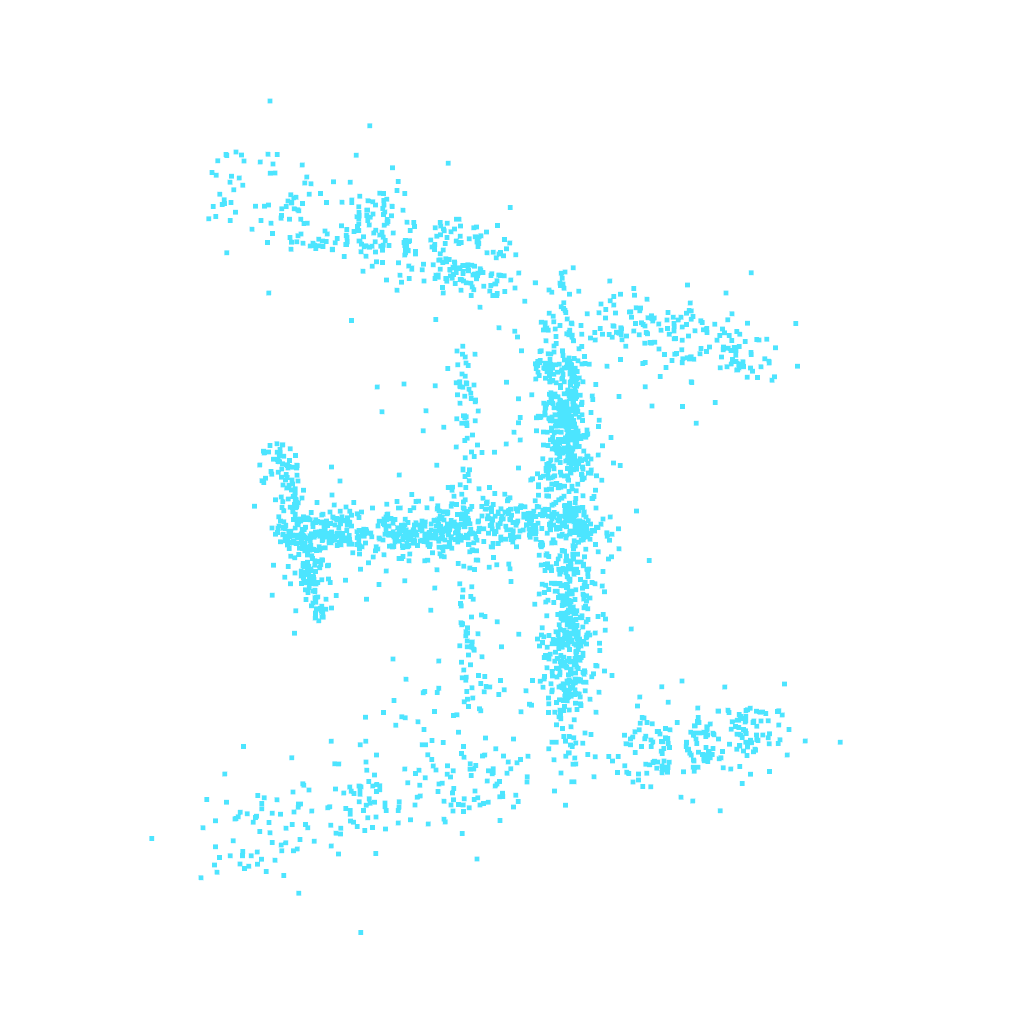}%
                \includegraphics[width=0.5\linewidth]{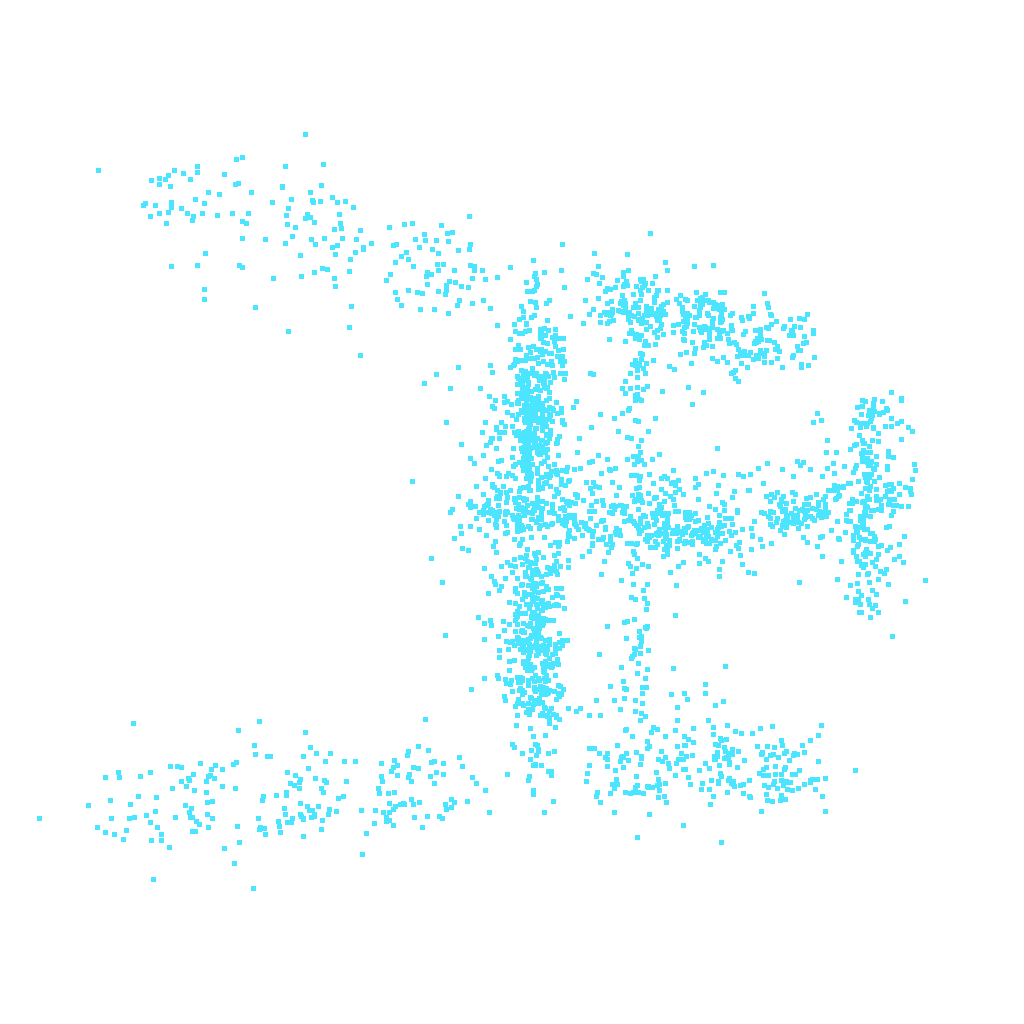}
                \subcaption*{Ours}
            \end{minipage}%
        \end{minipage}
    \end{minipage}
    
    % \vspace{2mm}
    {\raggedright (b) Novel View Synthesis Results \par}
    % \vspace{1mm}
        \begin{minipage}{\textwidth}
            \includegraphics[width=\linewidth]{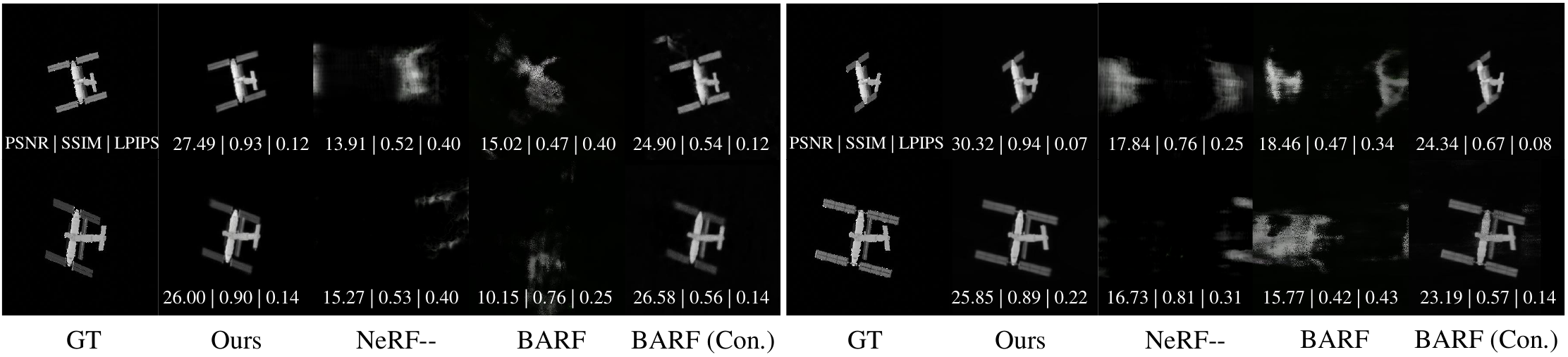}%
        \end{minipage}

    \caption{\textbf{Comparison of 3D reconstruction results on synthetic data using our method versus NeRF-based baselines.} (a) Reconstructed 3D point clouds from our method compared with the ground truth. (b) Novel view synthesis results from our method, NeRF-based baselines, and the ground truth. Results from two simulation datasets (1 \& 2) are presented. Our approach accurately recovers the satellite's 3D geometry, as well as produces clearer structures and higher SNR in novel views. In contrast, NeRF-based methods suffer from overfitting, resulting in pronounced artifacts and noise that diminish the quality of the reconstructed views.
    % Our results for 3D point cloud reconstruction (Including the sparse point cloud obtained using the modified Structure from Motion (SfM) and the final reconstructed point cloud), as well as novel view synthesis from various methods, demonstrate that our reconstruction closely aligns with the distribution of the real point clouds. Furthermore, our approach achieves a clear structural representation and a high signal-to-noise ratio in novel view synthesis. In contrast, NeRF-based methods exhibit signs of overfitting, leading to pronounced artifacts and noise in novel views, rendering them incapable of producing meaningful results.
    }
    \label{fig:compare_simu}%
    \end{figure*}

\subsubsection{Robust Pose Initialization}
From the 140 processed images, we select every tenth frame (e.g., 1st, 10th, 20th) for the training set and reserve the intermediate frames for validating the reconstruction.
We first compare the pose estimation accuracy of our method against existing techniques, including feature-based SfM methods—such as Colmap\cite{schonberger2016structure}, hierarchical localization\cite{sarlin2019coarse}, and detector-free SfM\cite{he2024detector}—as well as several feed-forward deep learning approaches, including DUSt3R\cite{wang2024dust3r} and RayDiffusion\cite{zhang2024cameras}. Fig.~\ref{fig:pose} and Table~\ref{tab:pose} compare the estimated poses to the ground truth (GT) on the training set. Feature-based SfM methods accumulate noise, which leads to inaccurate keypoint matching and causes bundle adjustment (BA) to fail to converge; consequently, no reasonable pose estimation can be reported. Similarly, deep learning methods struggle with extreme focal lengths, structural symmetry, and poor textures, resulting in incorrect depth estimation and subsequent pose errors. Other approaches, such as PoseDiffusion\cite{wang2023posediffusion}, MV-DUSt3R+\cite{tang2024mv}, and Colmap-free 3DGS\cite{fu2024colmap}, also exhibit substantial pose estimation errors.
In contrast, our pose initialization method achieves an ATE of only 2.61, with the estimated pose distribution closely aligning with the ground truth. As explained in Sec.\ref{subsubsec:pose_init}, our superior results stem from three key factors: (1) a robust orthographic projection model; (2) the use of RANSAC algorithms to eliminate erroneous matches; and (3) a strategy that leverages only the matching relationships between adjacent frames to relax the convergence conditions. These design choices enable our method to overcome the limitations of traditional SfM and deep learning approaches, significantly enhancing reliability for pose estimation of texture-poor space objects.

\begin{figure*}[!h]
    \centering
    % 上半部分：2行3列
    % \vspace{2mm}
    {\raggedright (a) Ours 3D Point Cloud Reconstruction Results \par}
    % \vspace{1mm}
    \noindent
    \begin{minipage}{0.49\textwidth}
    \centering
        China's Space Station (CSS)
        \end{minipage}%
    \hfill
    \begin{minipage}{0.49\textwidth}
        \centering
        International Space Station (ISS)
    \end{minipage}
    % \vspace{1mm}
    \begin{minipage}{\textwidth}
        \begin{minipage}{0.5\textwidth}
            \includegraphics[width=0.33\linewidth]{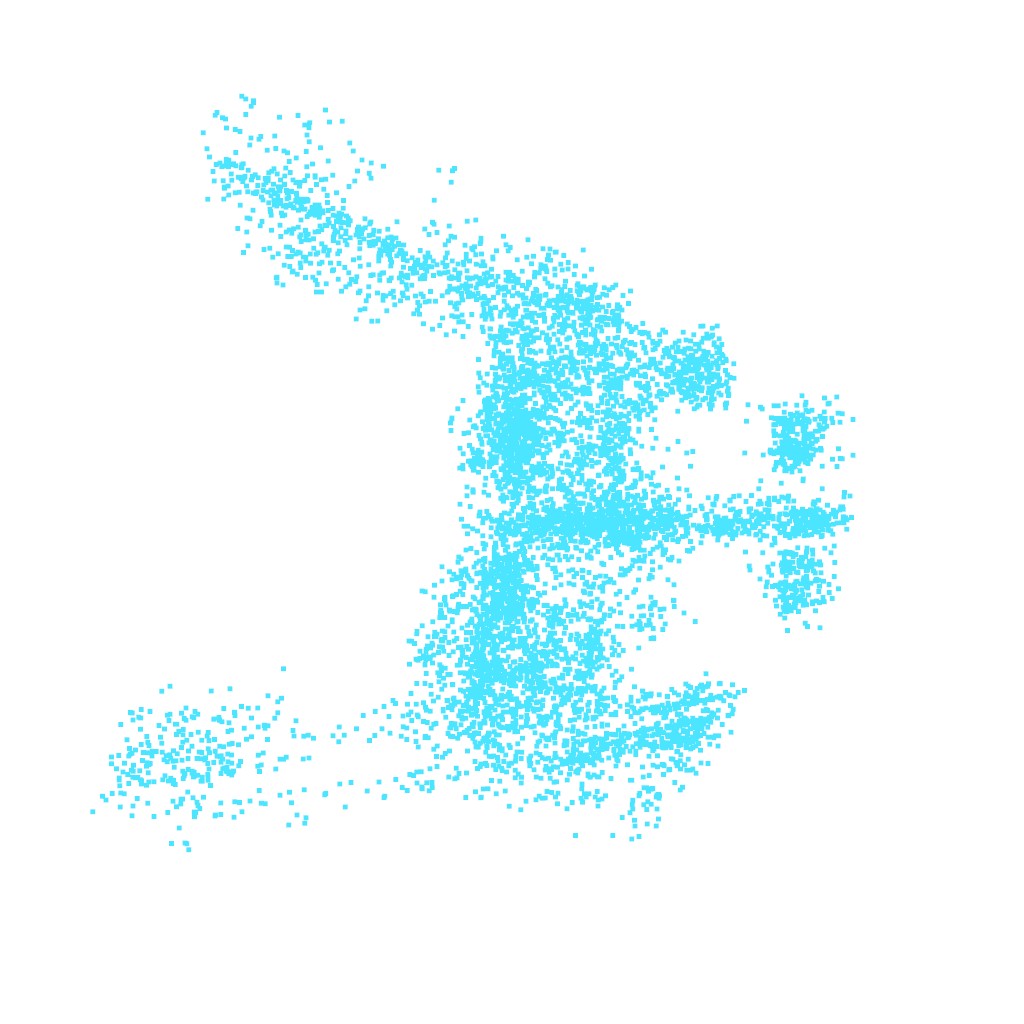}%
            \includegraphics[width=0.33\linewidth]{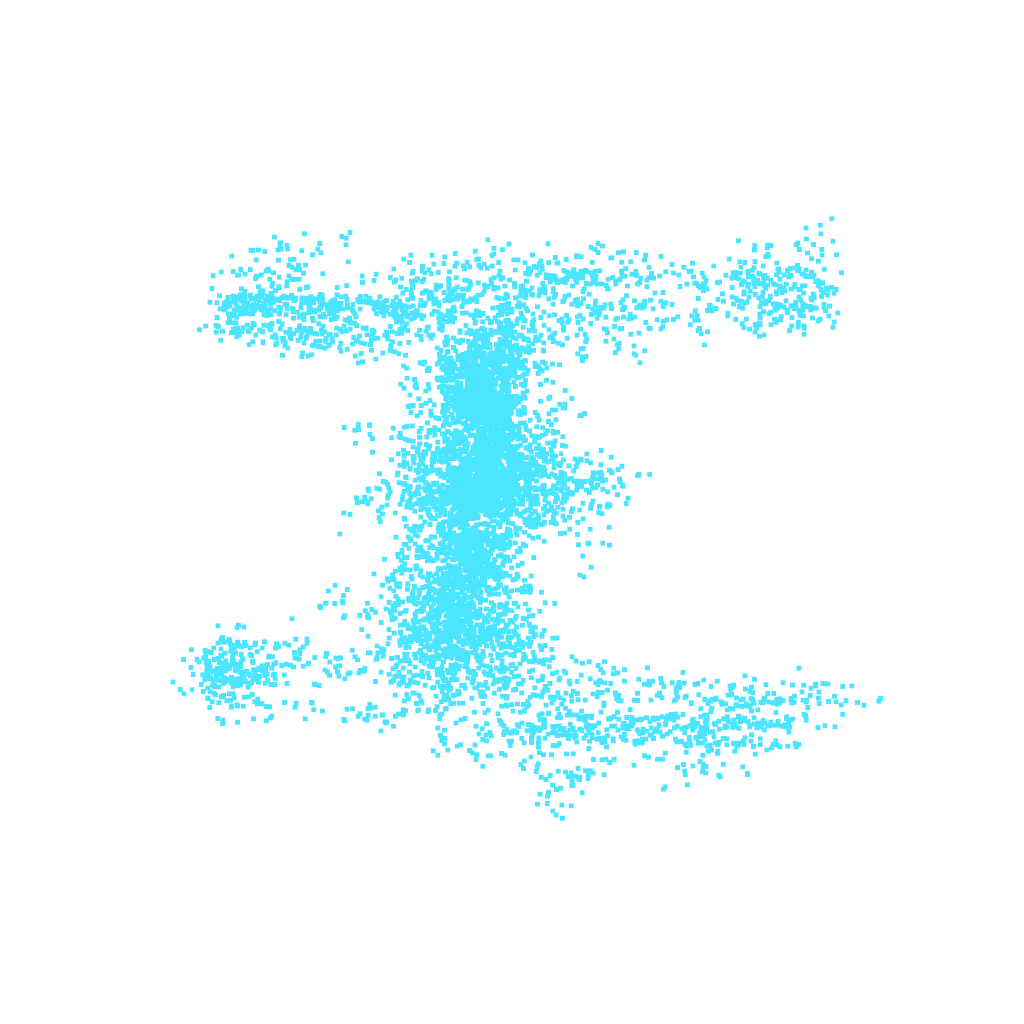}%
            \includegraphics[width=0.33\linewidth]{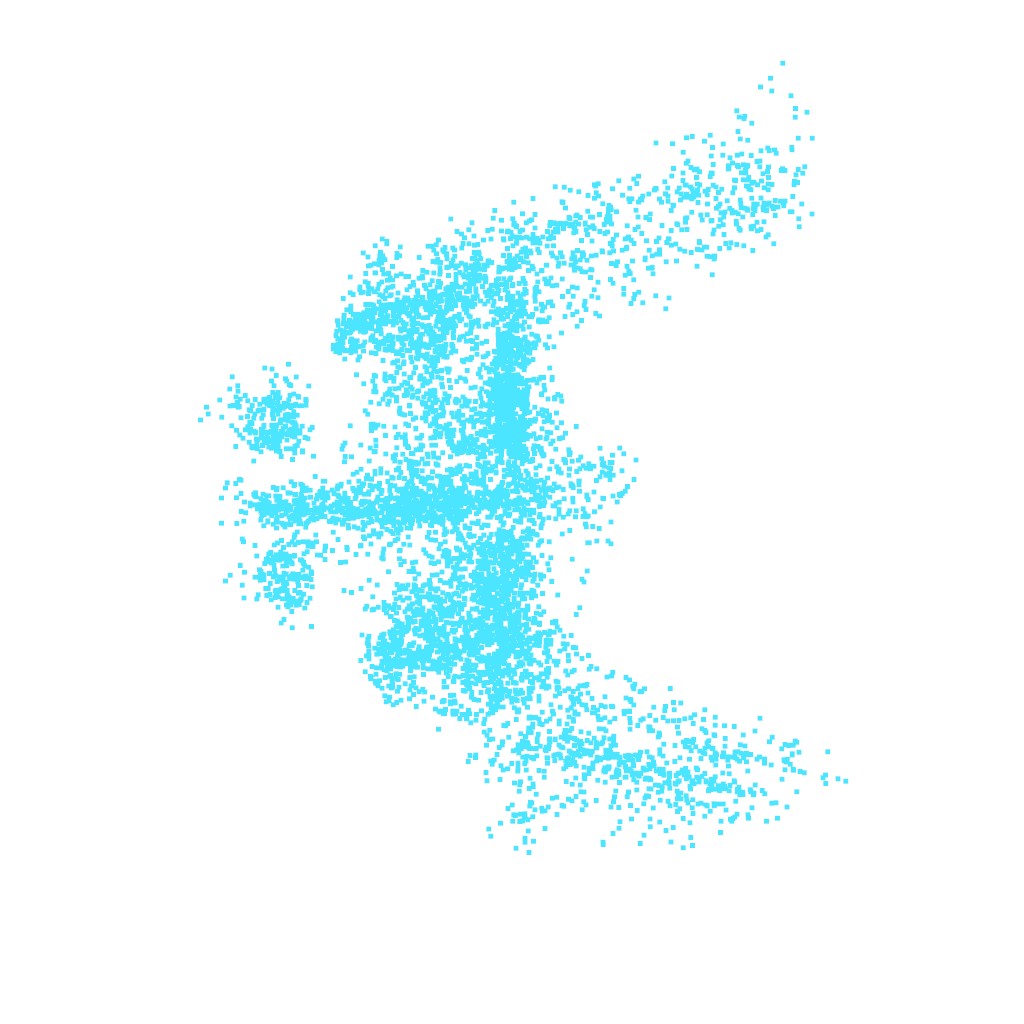}%
            % \subcaption{China's Space Station (CSS)}%
        \end{minipage}%
        \begin{minipage}{0.5\textwidth}
            \includegraphics[width=0.33\linewidth]{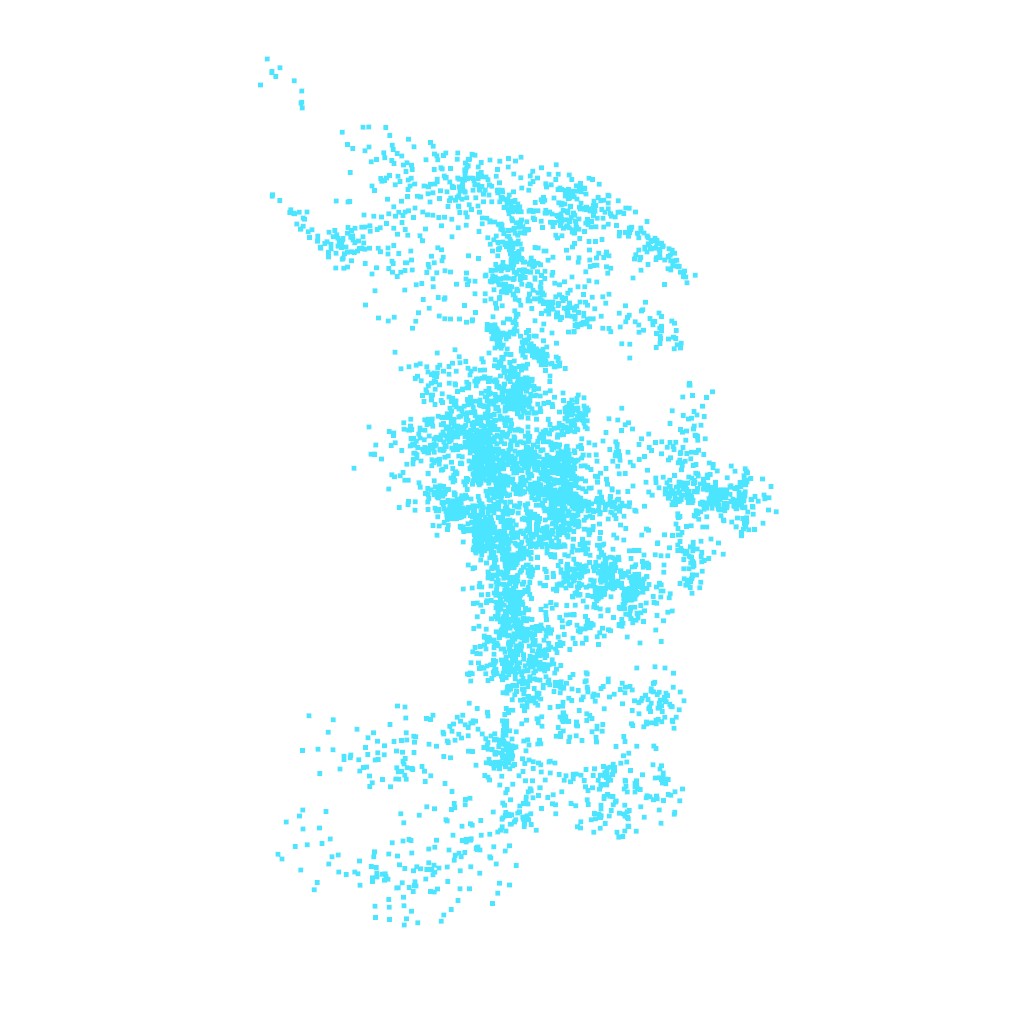}%
            \includegraphics[width=0.33\linewidth]{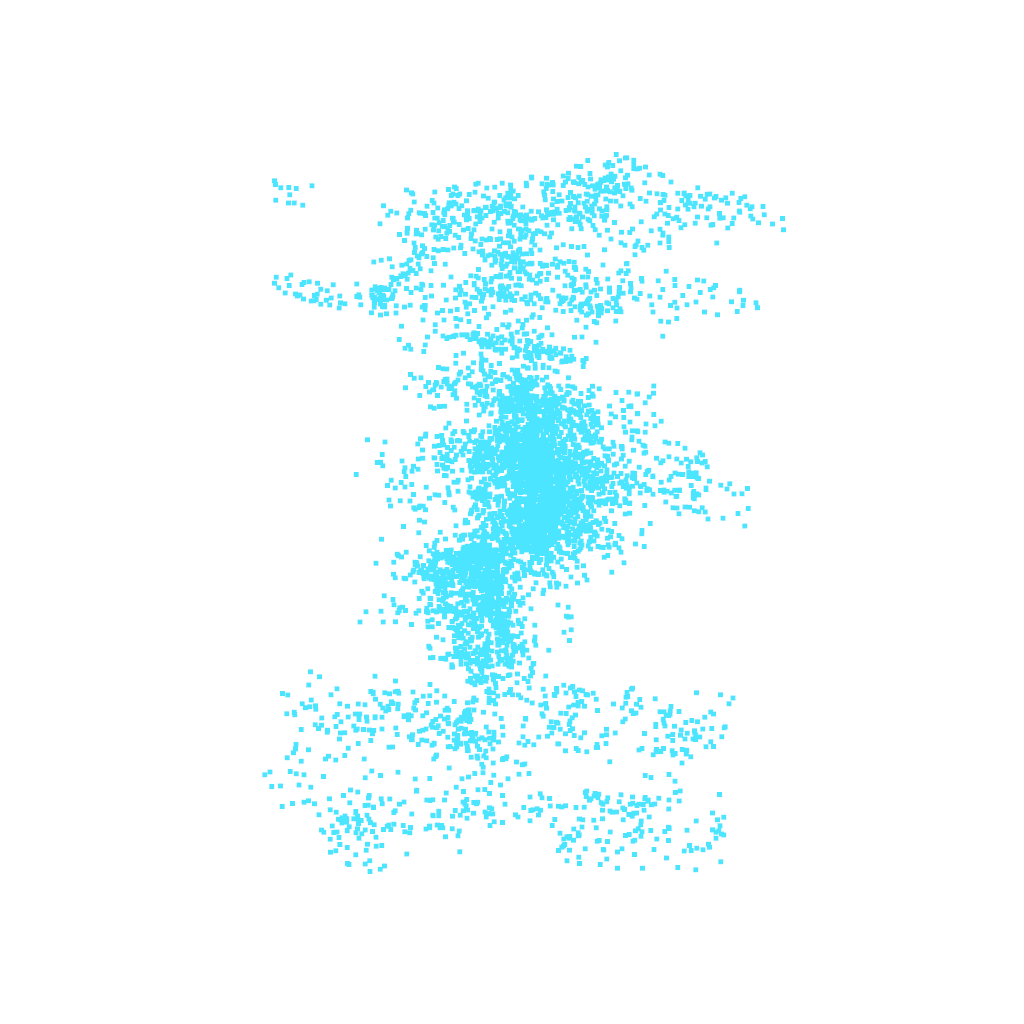}%
            \includegraphics[width=0.33\linewidth]{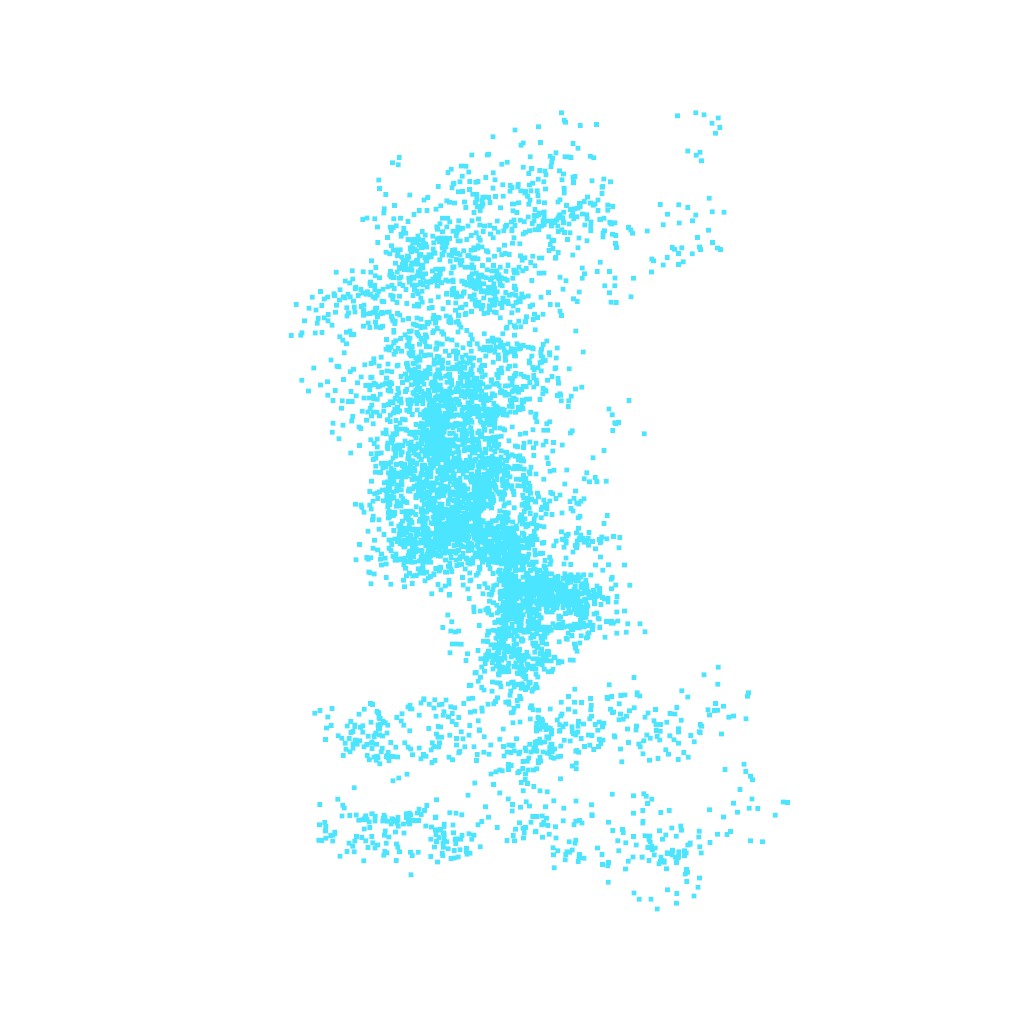}%
            % \subcaption{International Space Station (ISS)}%
        \end{minipage}
        \vspace{-3mm}
    \end{minipage}
    {\raggedright (b) Novel View Synthesis Results \par}
    \begin{minipage}{\textwidth}
        \includegraphics[width=\linewidth]{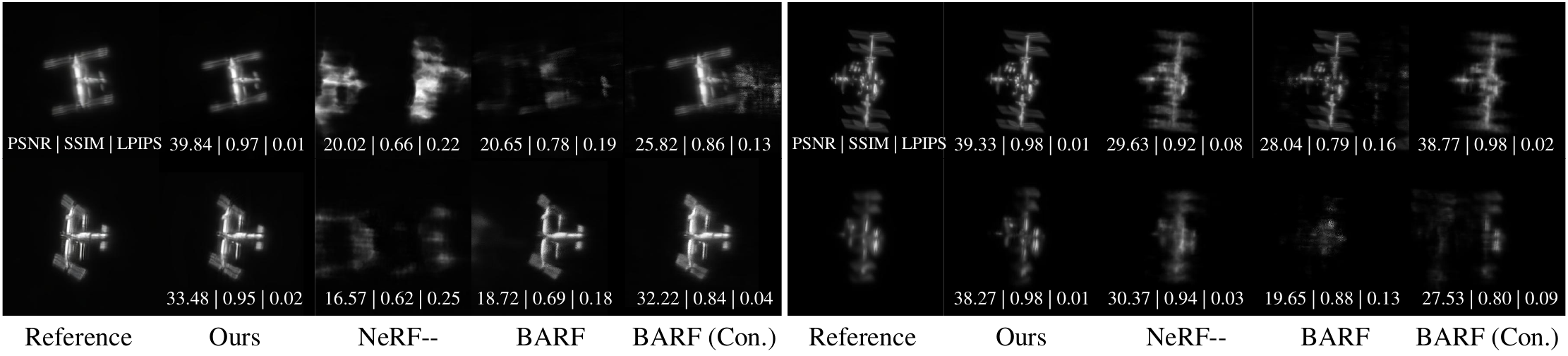}%
    \end{minipage}

    \caption{\textbf{Comparison of 3D reconstruction results on real observations using our method versus NeRF-based baselines.} (a) Reconstructed 3D point clouds produced by our method from multiple viewpoints. (b) Novel view synthesis results from our method and NeRF-based baselines. Results from two on-sky observations, CSS and ISS, are presented. Our approach accurately recovers the satellite's 3D geometry, as well as produces clearer structures and higher SNR in novel views than NeRF-based methods, which use reserved pre-processed images as references.
    % Our reconstructed point clouds and novel view synthesis results from various methods. Our point clouds exhibit clear structure and align with factual representations. In terms of novel view synthesis, our results showcase clear structural representation and high signal-to-noise ratio. In contrast, NeRF-based methods suffer from significant artifacts, blurriness, and noise in novel views, ultimately failing to produce meaningful results.
    }
    \label{fig:compare_real}
\end{figure*}

\subsubsection{Joint 3D Reconstruction and Pose Refinement}
We then compare our pipeline with two NeRF-based baseline methods, NeRF$--$ and BARF, that support joint 3D reconstruction and pose estimation. To ensure a fair comparison, we use our customized pose initialization for all baselines; otherwise, neither NeRF$--$ nor BARF exhibits generalization ability. In addition, traditional NeRF-based methods uniformly sample the space between the cameras and the object (illustrated by the light blue dots in Fig.~\ref{fig:sample}). This approach fails in our sparse-view, long-distance observational scenario because of overfitting in the excessively large sampling space. Consequently, we explore a BARF variant (named BARF (Con.)) that restricts the sampling region to the vicinity of the initial object distance predicted by our modified SfM (indicated by the gray dots in Fig.~\ref{fig:sample}). This adaptive strategy significantly reduces overfitting and improves reconstruction accuracy.

Fig.~\ref{fig:compare_simu} compares the reconstructed point clouds (appearance and shape attributes of 3D Gaussians neglected) and novel views from our method with those produced by baseline techniques. Although NeRF-based methods have been carefully tuned for this task, they still struggle to generalize to unseen viewpoints. In contrast, our GS-based approach employs controlled point growth to focus sampling within the object region, accurately capturing fine structural details. This advantage is confirmed by superior SSIM, PSNR, and LPIPS metrics (Table~\ref{tab:result} and Fig.~\ref{fig:metrics}). In the meantime, our method delivers more accurate pose estimates after the joint optimization of poses and the 3D object (Table~\ref{tab:pose}).

% Our method achieves the lowest Absolute Trajectory Error (ATE) after joint optimization of poses and the 3D scene, as shown in Table~\ref{tab:pose}. This indicates not only more accurate camera pose estimates throughout the reconstruction process, but also a more precise 3D reconstruction, as improved pose accuracy enhances scene recovery.

To further quantify the precision of our reconstructions, we measure the Chamfer distance (CD) between the reconstructed models and the ground-truth point clouds. As illustrated in Table~\ref{tab:result}, our method demonstrates significantly lower CD values, showcasing its ability to accurately capture 3D geometries. As visualized by the point clouds and the novel views in Fig.~\ref{fig:compare_simu}, our method accurately recovers the 3D structure of the satellite, while NeRF-based methods fail to extract any structural information.

Our method also significantly reduces training time. A 3D satellite model is reconstructed in approximately 4 minutes on a single NVIDIA RTX 4090 using our method (2 minutes for GS training and 2 minutes for BnB pose refinement), much faster than NeRF$--$'s 1.5 hours and BARF's over 4 hours.

\begin{figure}[!h]
    \centering
    \includegraphics[width=0.4\textwidth]{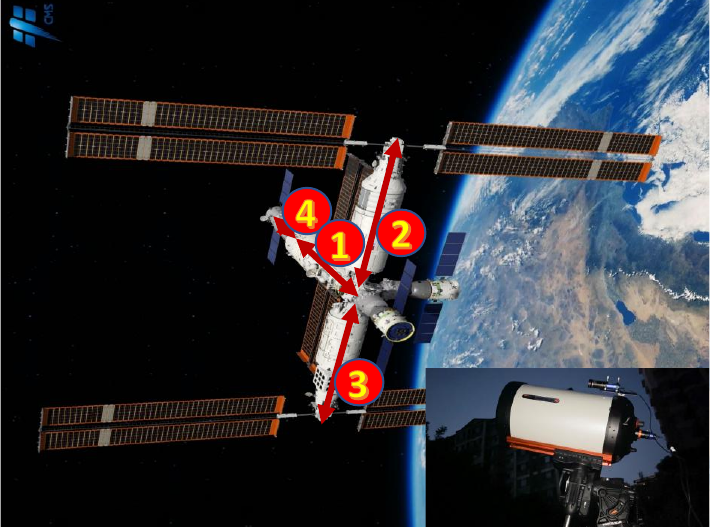}
    \caption{\textbf{The Chinese Space Station (CSS) and the imaging instrument.} On the left, the official schematic of the CSS, featuring four primary modules: 1 - Tianhe Core Module, 2 - Mengtian Laboratory Module, 3 - Wentian Laboratory Module, and 4 - Tianzhou-6 Cargo Spacecraft. In the lower right corner, the Celestron C14HD telescope paired with a QHY5III678M planetary camera.}
    \label{fig:css}
    \vspace{-8pt}
\end{figure}

\subsection{On-Sky Results}
We apply our algorithm to real observations of China's Tiangong Space Station (CSS) and the ISS. Since both low-Earth orbit satellites have well-documented geometric specifications, we are able to assess our method's accuracy by comparing the reconstructions with publicly available information.

\setlength{\tabcolsep}{4pt}
\begin{table}[!h]
\centering
\caption{\textbf{Metrology of the China's Space Station modules and their solar wings.}}
\label{tab:measure}
\begin{tabular}{llcc}
\toprule
\textbf{Category} & \textbf{Module} & \textbf{Measured} & \textbf{GT\cite{wiki:tiangong}} \\ 
\midrule
\multirow{4}{*}{\textbf{Length}} 
  & Tianhe Core Module   & 16.11 & 16.6 \\
  & Mengtian Laboratory  & 17.30 & 17.88 \\
  & Wentian Laboratory   & 17.30 & 17.88 \\
  & Tianzhou-6           & 9.71  & 10.6 \\ 
\midrule
\multirow{4}{*}{\textbf{Diameter}} 
  & Tianhe Core Module   & 5.65  & 4.2 \\
  & Mengtian Laboratory  & 5.50  & 4.2 \\
  & Wentian Laboratory   & 5.63  & 4.2 \\
  & Tianzhou-6           & 3.30  & 3.35 \\ 
\midrule
\multirow{4}{*}{\textbf{Solar wing length}} 
  & Tianhe Core Module   & 25.44 & $\times$ \\
  & Mengtian Laboratory  & 55.62 & 55 \\
  & Wentian Laboratory   & 55.62 & 55 \\
  & Tianzhou-6           & 4.89  & $\times$ \\ 
\midrule
\multirow{4}{*}{\textbf{Solar wing angle}} 
  & Tianhe Core Module   & 0.78  & $\times$ \\
  & Mengtian Laboratory  & 63.01 & $\times$ \\
  & Wentian Laboratory   & 68.71 & $\times$ \\
  & Tianzhou-6           & 3.22  & $\times$ \\ 
\bottomrule
\end{tabular}
\vspace{-8pt}
\end{table}

% \subsubsection{Observations of CSS and ISS}
% The CSS was observed on September 15, 2023, from 20:48:30 to 20:51:17 UTC near Miyun Reservoir, Beijing, with a closest approach of 389.4 km and a peak elevation of 78.3°. The ISS was observed on August 21, 2024, from 20:20:06 to 20:22:09 UTC in Yanqing District, Beijing, with a closest approach of 442 km and a peak elevation of 86°. For both observations, we used a Celestron C14HD telescope (0.35 m diameter, f/11) paired with a QHY5III678M camera (Fig.~\ref{fig:telescope}), achieving a spatial sampling rate of 0.106 arcsec/pixel (approximately 0.2 m/pixel). Our observations employed the "Fake Polar Axis Method" with a German equatorial mount (CGX-L), uniquely oriented to a precisely calculated azimuth relative to the satellite's track rather than true north. This innovative approach eliminates the need for meridian flips and minimizes field rotation, particularly in the center of the image sequence. Images were captured at 5 ms exposures, averaging 88 fps for CSS and 55 fps for ISS, resulting in datasets of 13,800 images for CSS and 6,770 for ISS, with resolutions of 360 and 640, respectively.

\begin{figure*}[!h]
    \centering
    % 上半部分：2行3列
    % \vspace{2mm}
    {\raggedright \hspace*{1em} (a) 3D Point Cloud Reconstruction Results \par}
    % \vspace{1mm}
    \begin{minipage}{\textwidth}
        \begin{minipage}{0.14\textwidth}
            \includegraphics[width=\linewidth]{pointcloud/simu3_gt.png}
            \subcaption{GT}
        \end{minipage}%
        \begin{minipage}{0.14\textwidth}
            \includegraphics[width=\linewidth]{pointcloud/simu3_ours}
            \subcaption{Ours}
        \end{minipage}%
        \begin{minipage}{0.14\textwidth}
            \includegraphics[width=\linewidth]{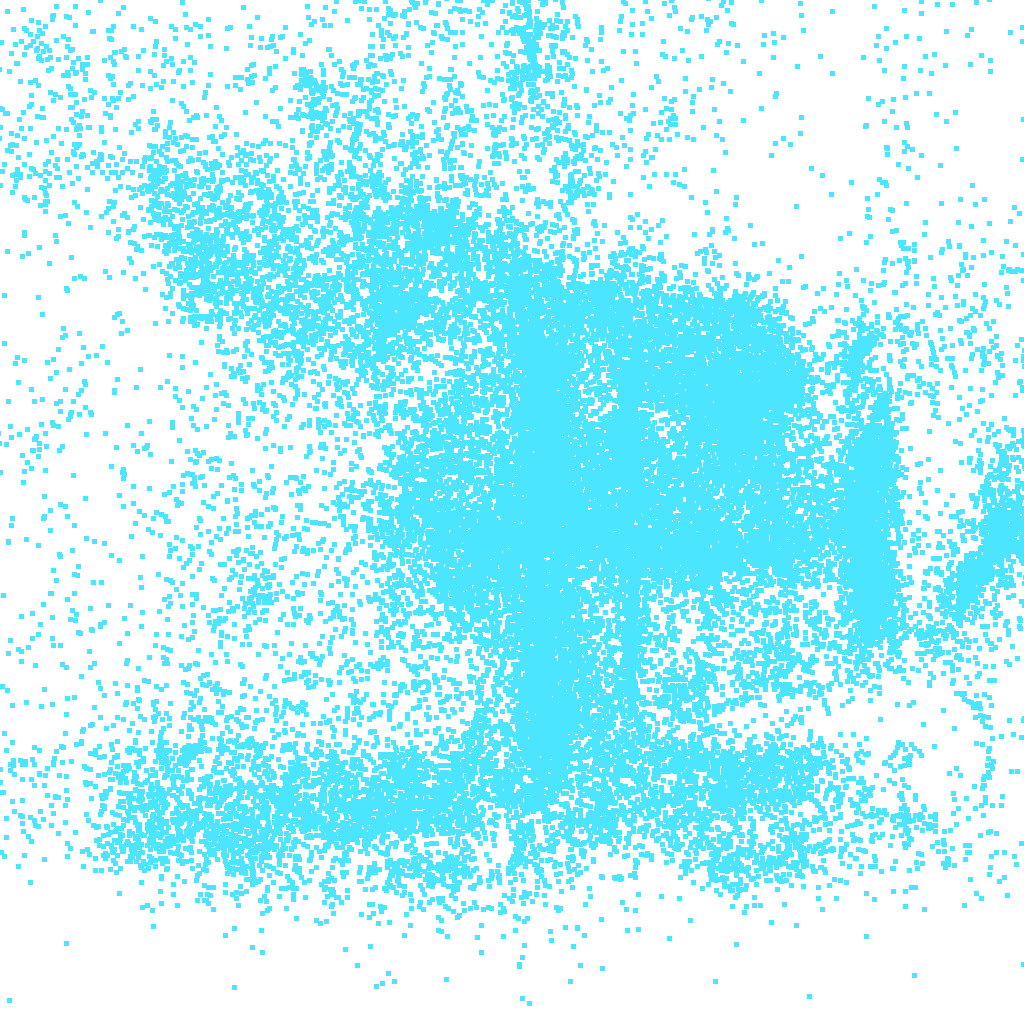}
            \subcaption{Original GS}
        \end{minipage}%
        \begin{minipage}{0.14\textwidth}
            \includegraphics[width=\linewidth]{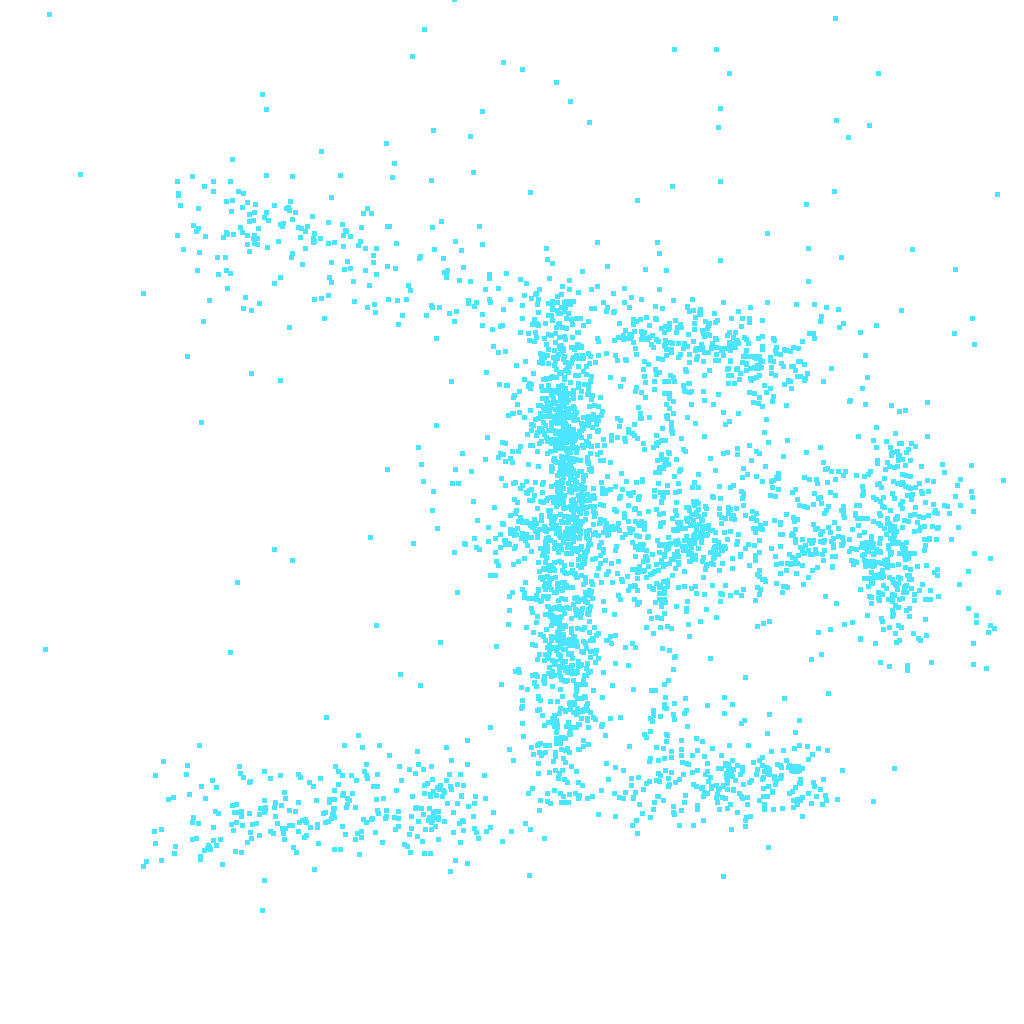}
            \subcaption{No Preprocessing}
        \end{minipage}%
        \begin{minipage}{0.14\textwidth}
            \includegraphics[width=\linewidth]{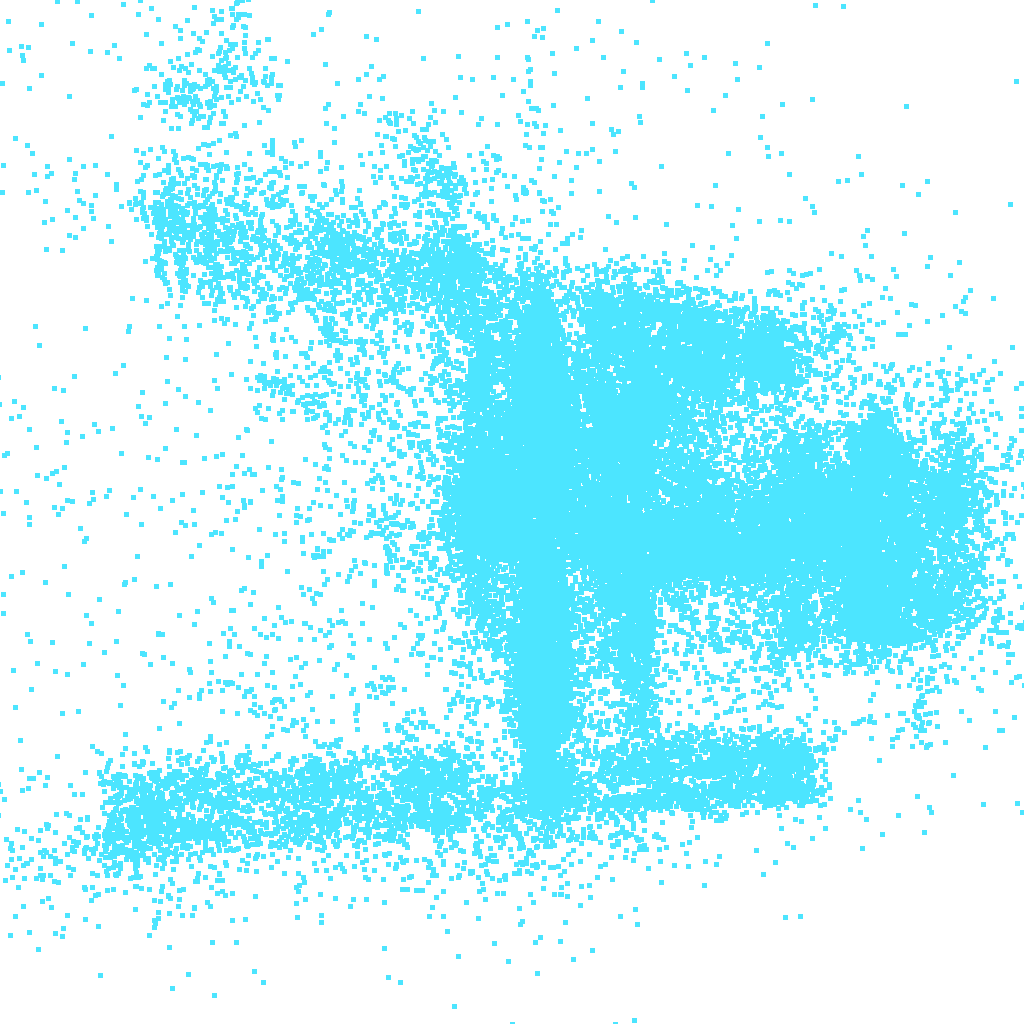}
            \subcaption{No Regulation}
        \end{minipage}%
        \begin{minipage}{0.14\textwidth}
            \includegraphics[width=\linewidth]{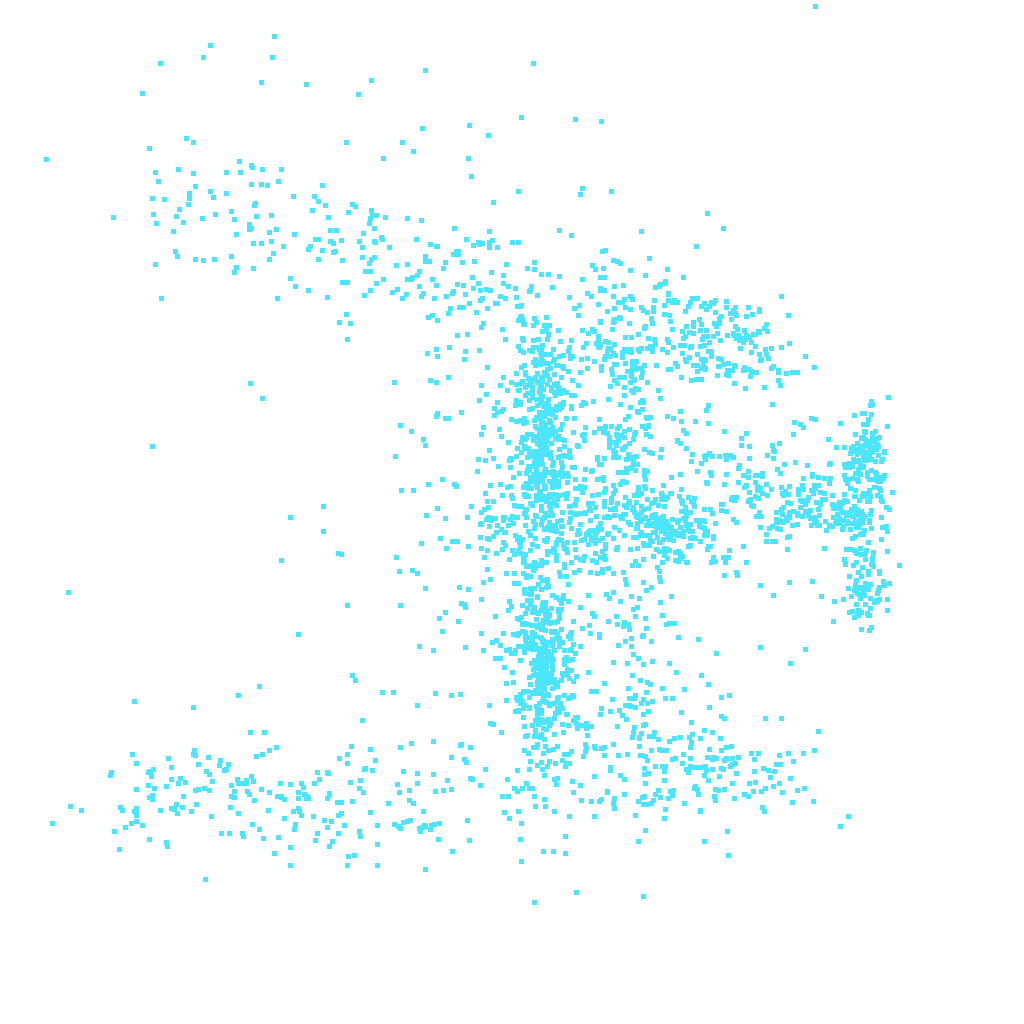}
            \subcaption{No BnB}
        \end{minipage}%
        \begin{minipage}{0.14\textwidth}
            \includegraphics[width=\linewidth]{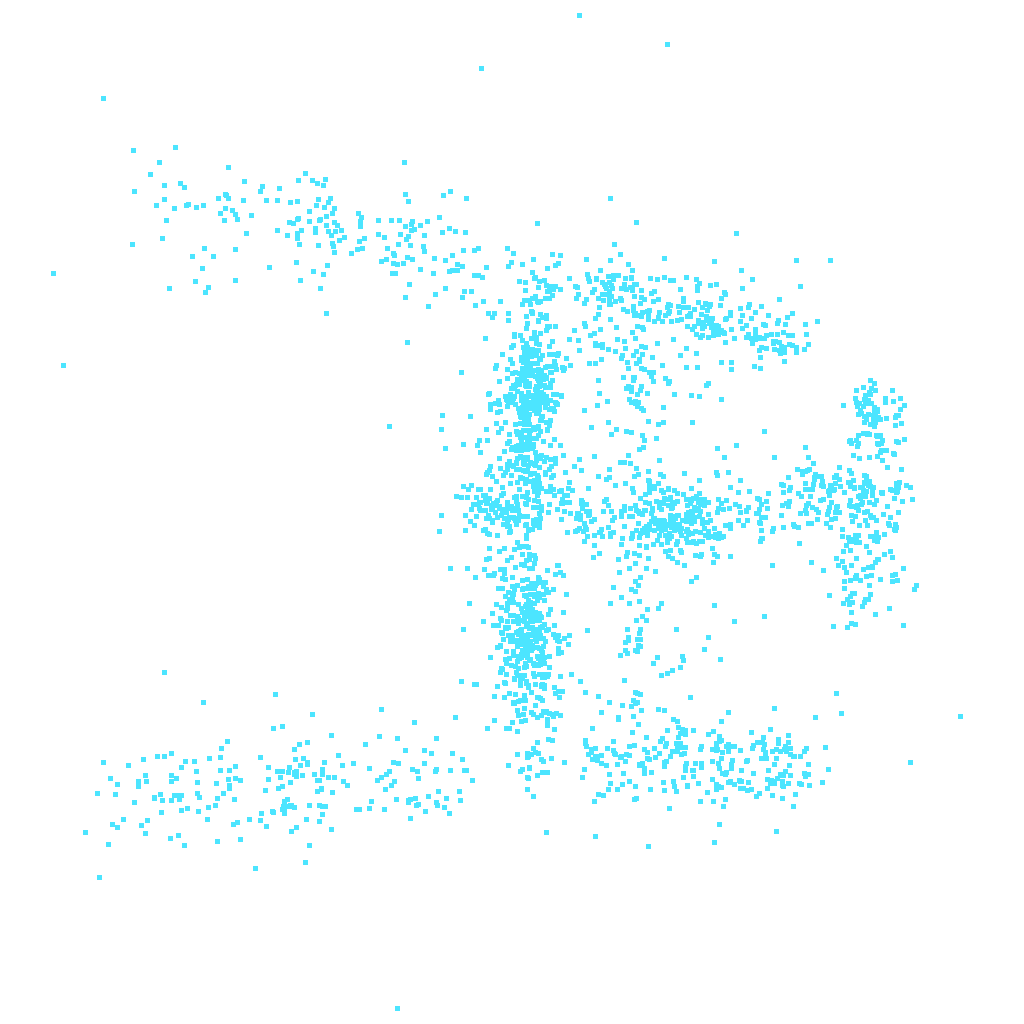}
            \subcaption{No Filtering}
        \end{minipage}
    \end{minipage}
    
    % \vspace{4mm}
    {\raggedright \hspace*{1em} (b) Novel View Synthesis results\par}
    % \vspace{1mm}
    \begin{minipage}{\textwidth}
        \centering
        \begin{minipage}{0.14\textwidth}
            \includegraphics[width=\linewidth]{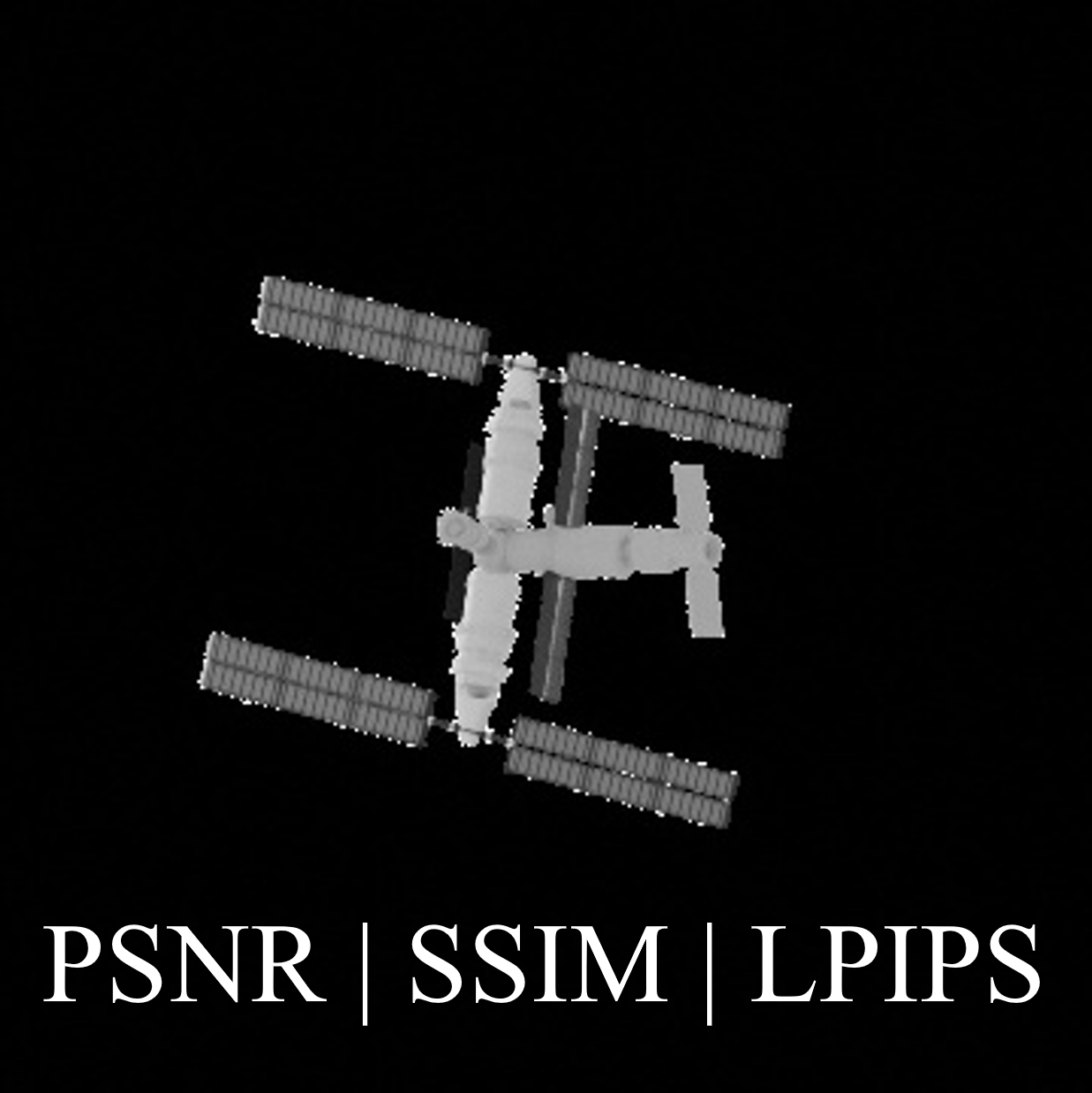}
        \end{minipage}%
        \begin{minipage}{0.14\textwidth}
            \includegraphics[width=\linewidth]{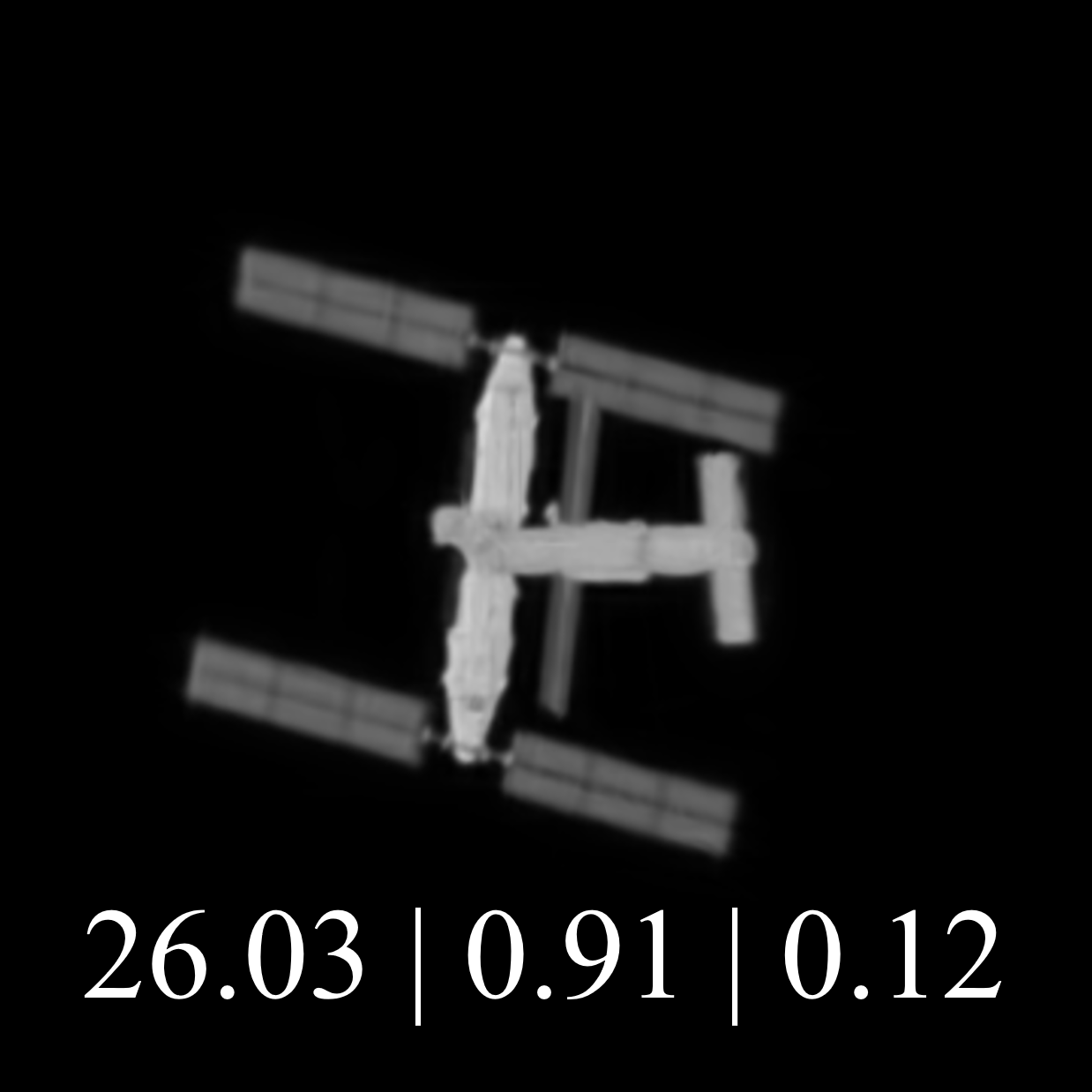}
        \end{minipage}%
        \begin{minipage}{0.14\textwidth}
            \includegraphics[width=\linewidth]{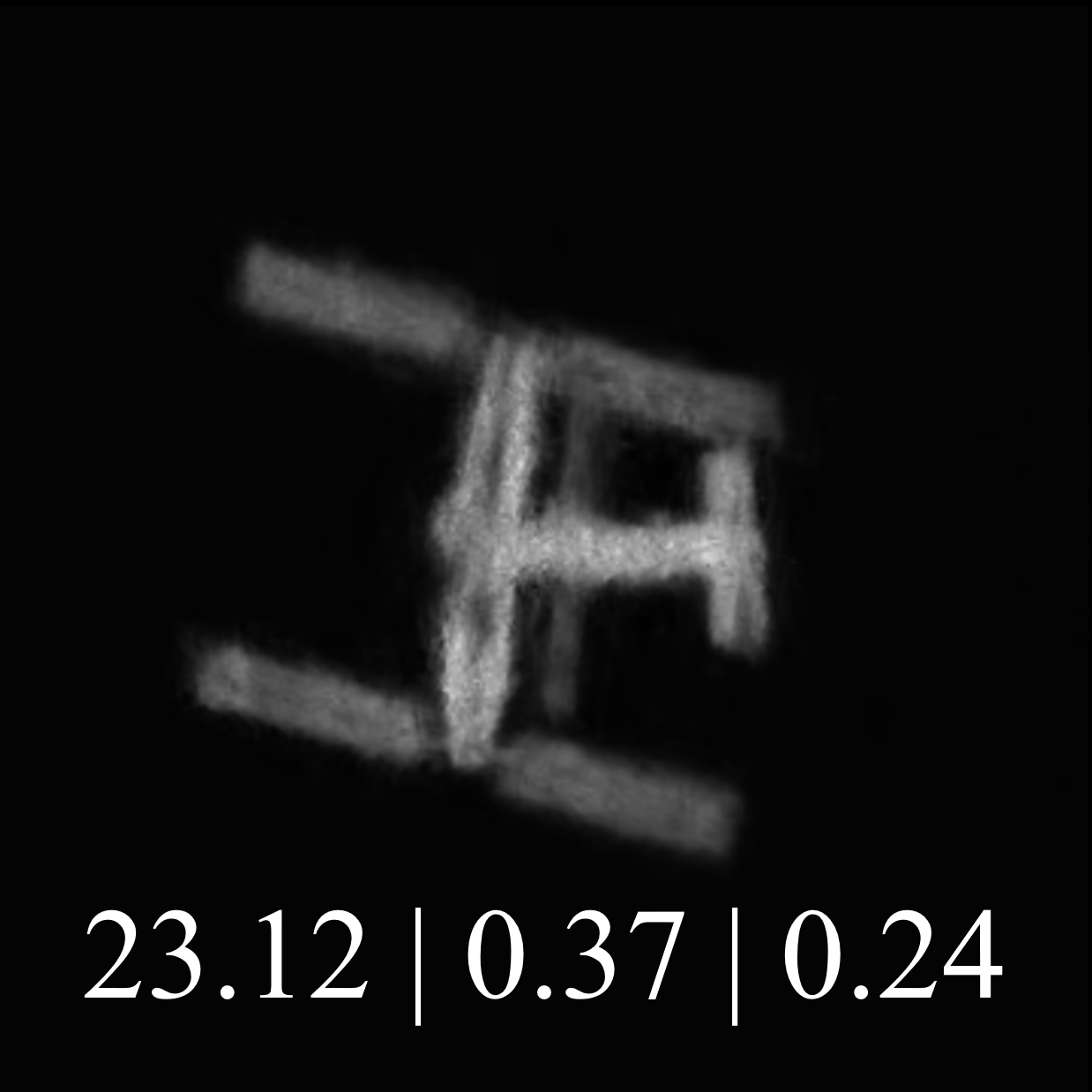}
        \end{minipage}%
        \begin{minipage}{0.14\textwidth}
            \includegraphics[width=\linewidth]{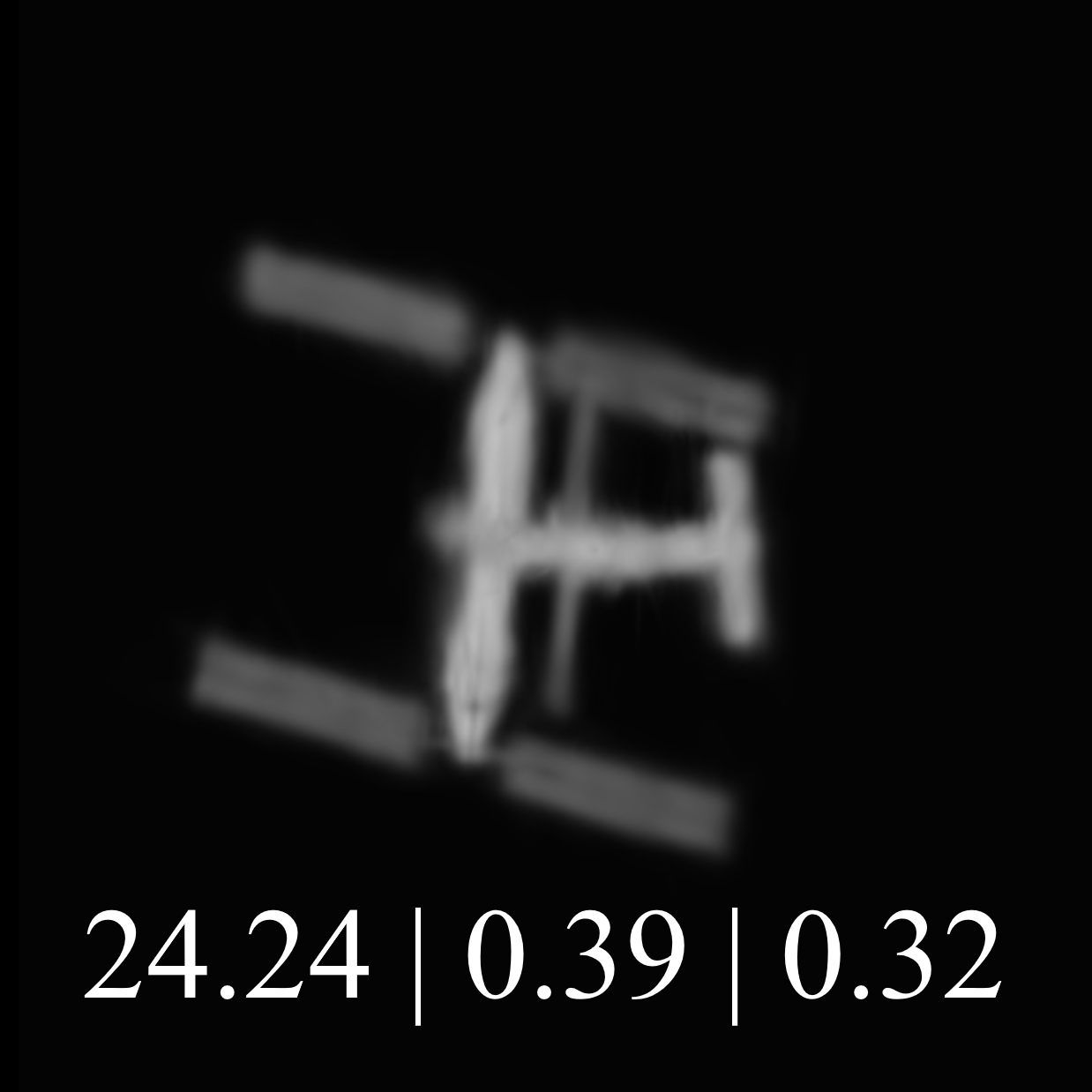}
        \end{minipage}%
        \begin{minipage}{0.14\textwidth}
            \includegraphics[width=\linewidth]{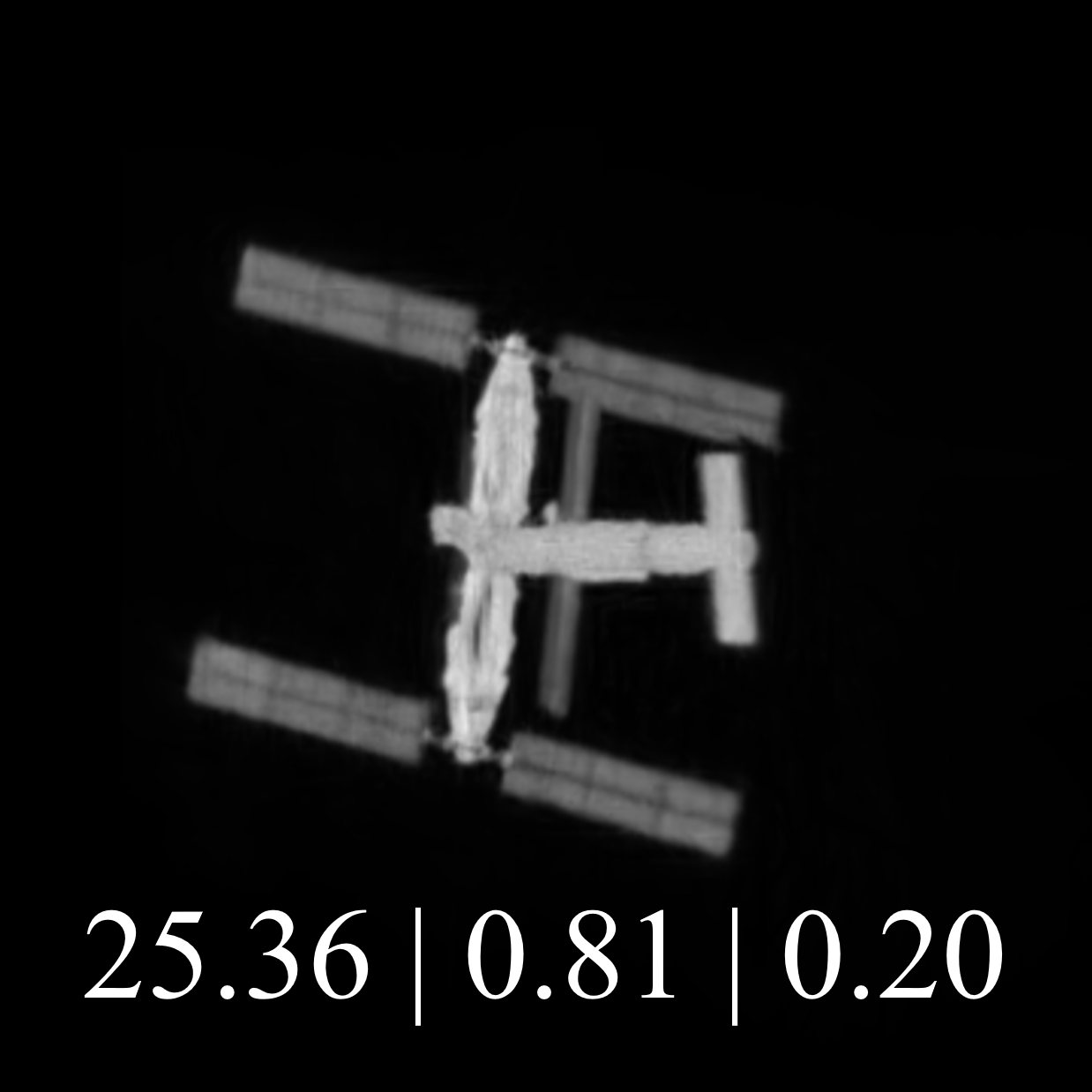}
        \end{minipage}%
        \begin{minipage}{0.14\textwidth}
            \includegraphics[width=\linewidth]{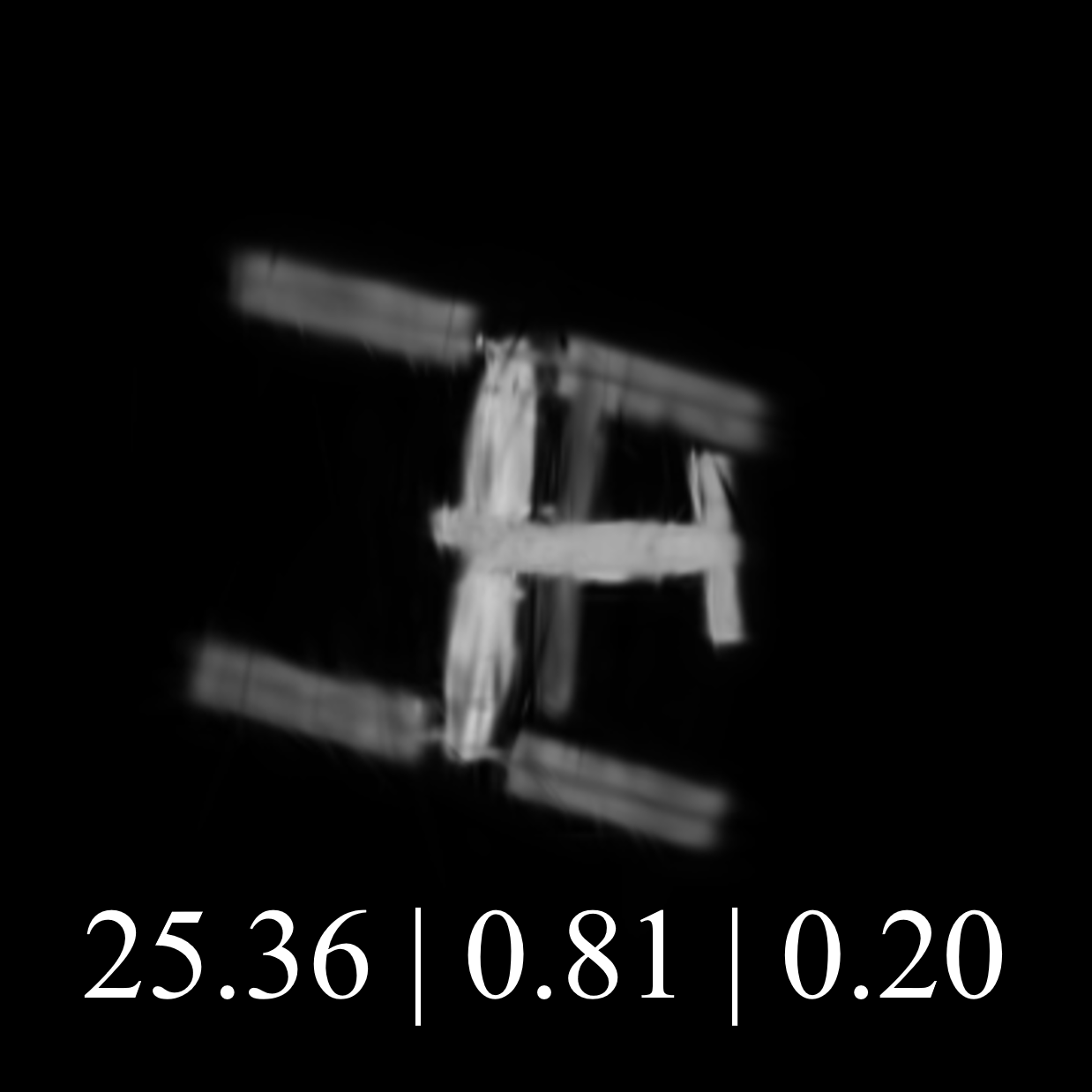}
        \end{minipage}%
        \begin{minipage}{0.14\textwidth}
            \includegraphics[width=\linewidth]{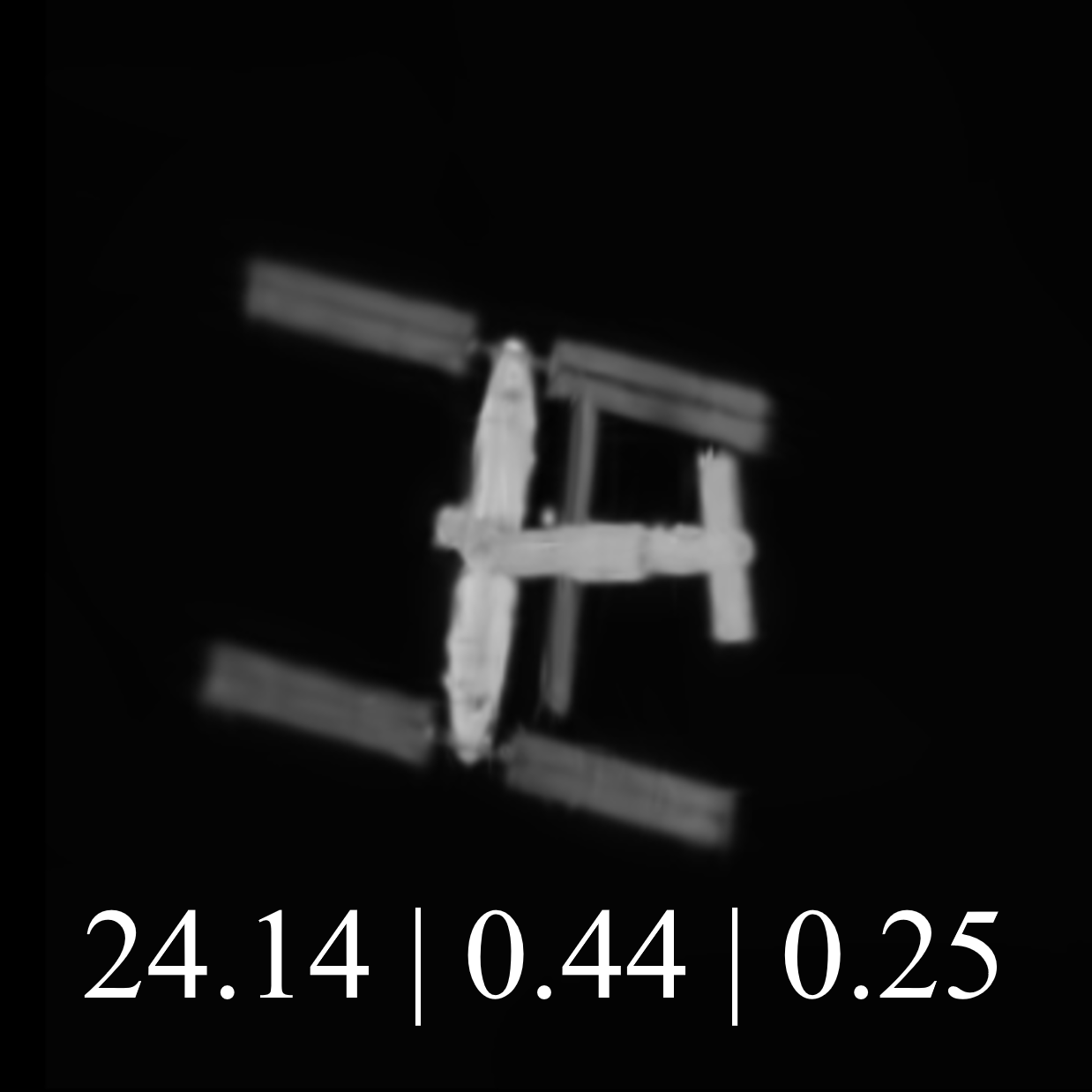}
        \end{minipage}

        \begin{minipage}{0.14\textwidth}
            \includegraphics[width=\linewidth]{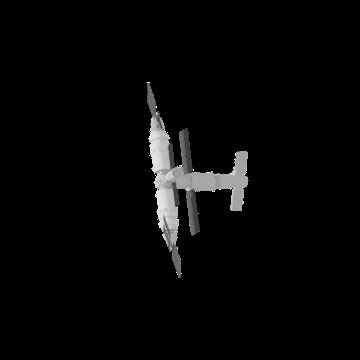}
            \subcaption{GT}
        \end{minipage}%
        \begin{minipage}{0.14\textwidth}
            \includegraphics[width=\linewidth]{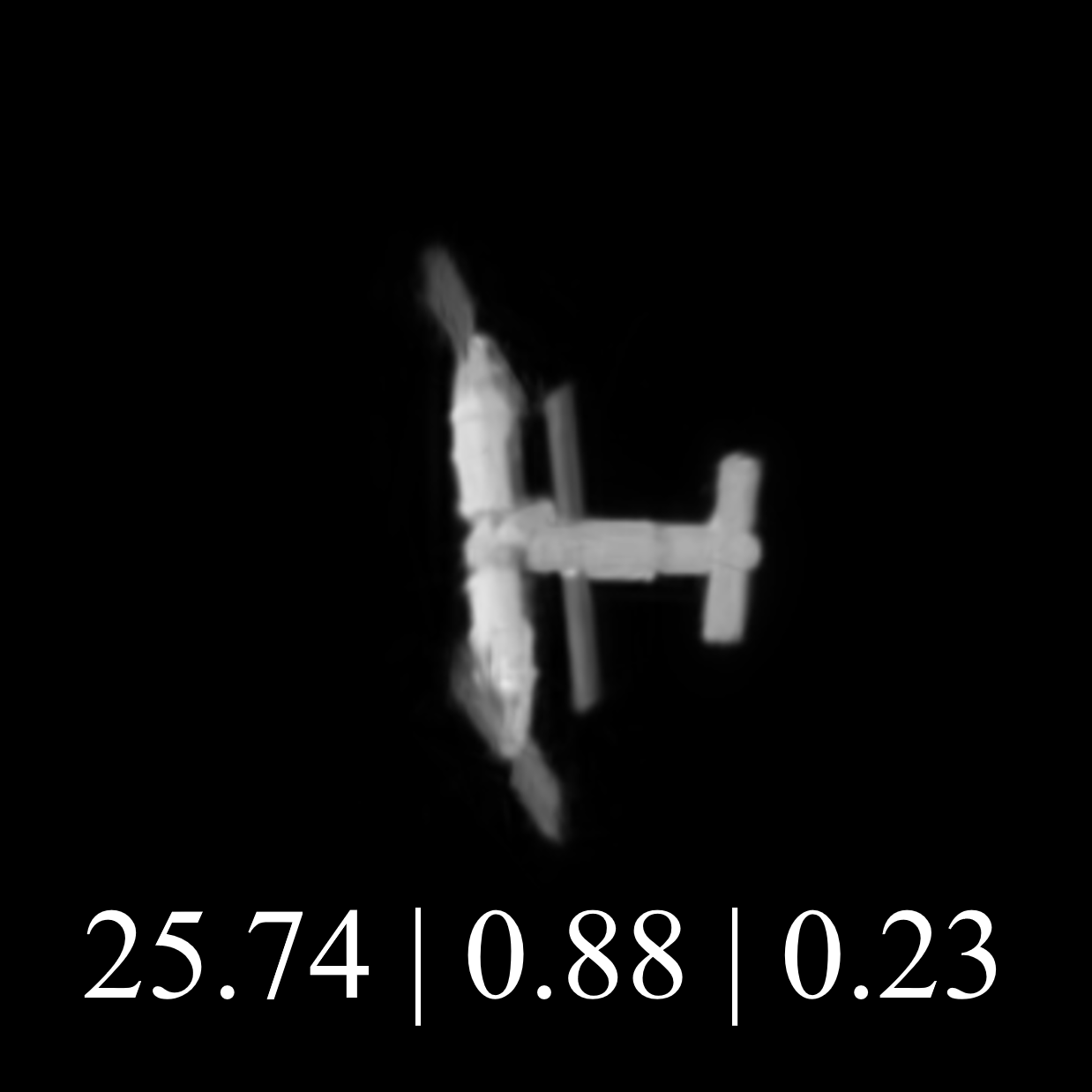}
            \subcaption{Ours}
        \end{minipage}%
        \begin{minipage}{0.14\textwidth}
            \includegraphics[width=\linewidth]{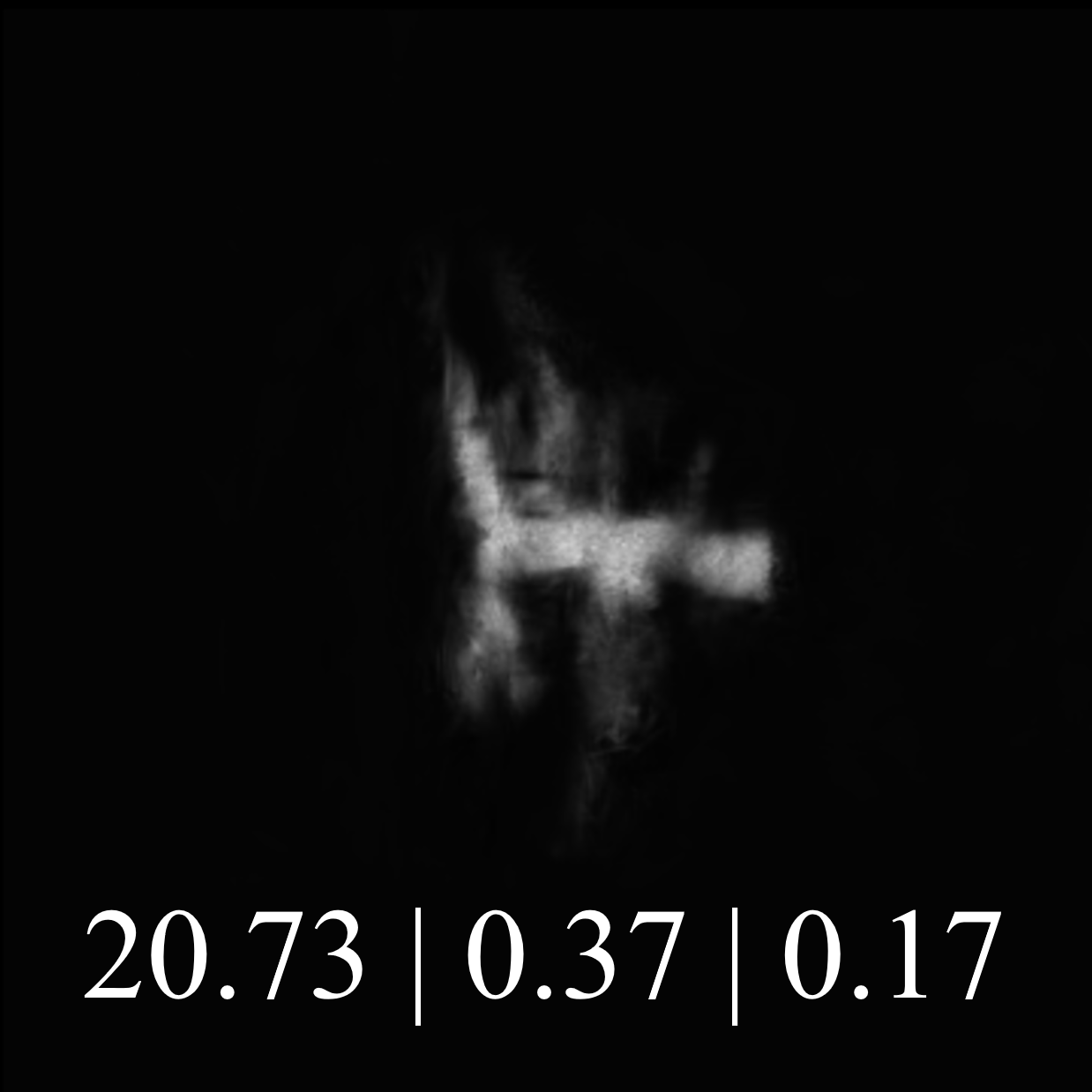}
            \subcaption{Original GS}
        \end{minipage}%
        \begin{minipage}{0.14\textwidth}
            \includegraphics[width=\linewidth]{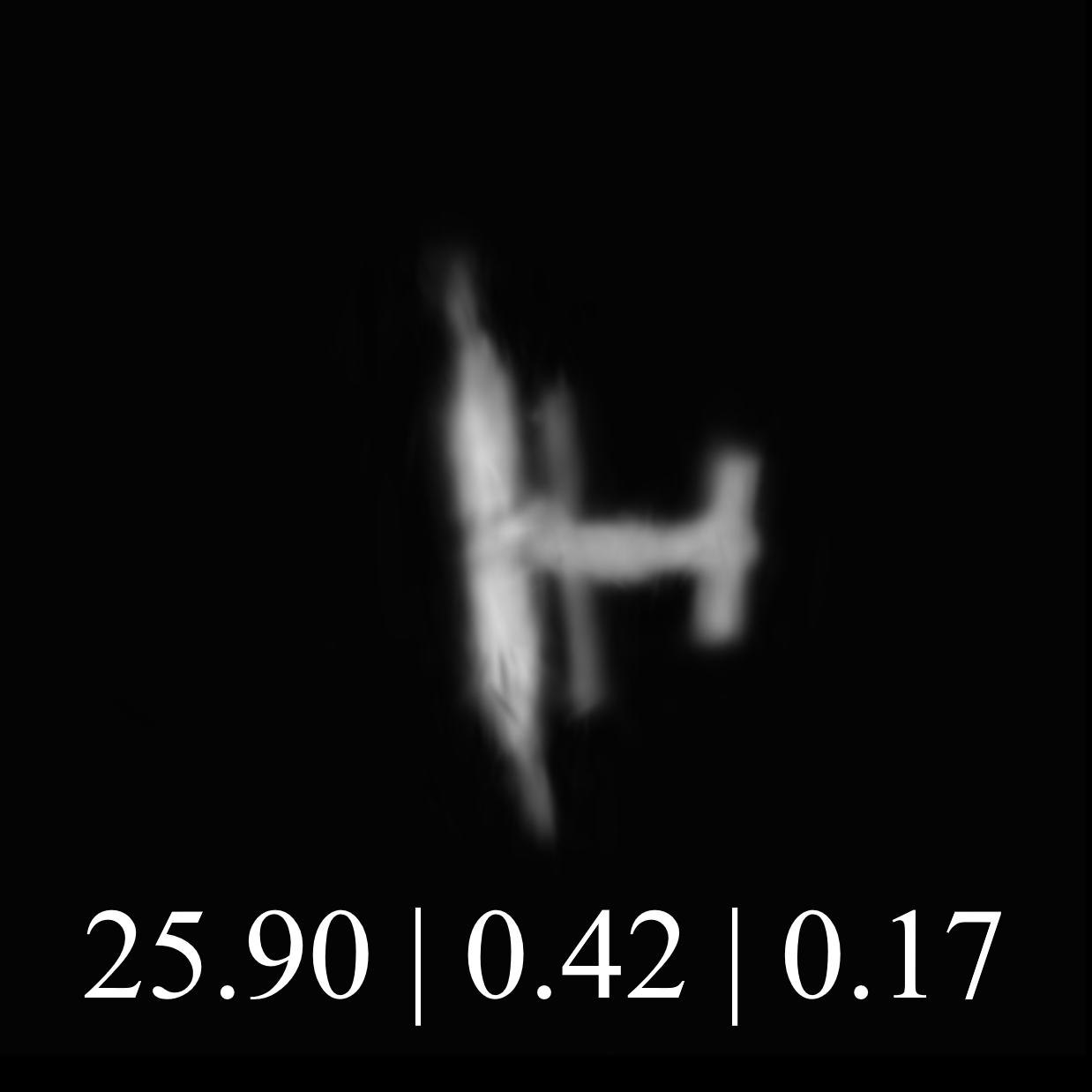}
            \subcaption{No Preprocessing}
        \end{minipage}%
        \begin{minipage}{0.14\textwidth}
            \includegraphics[width=\linewidth]{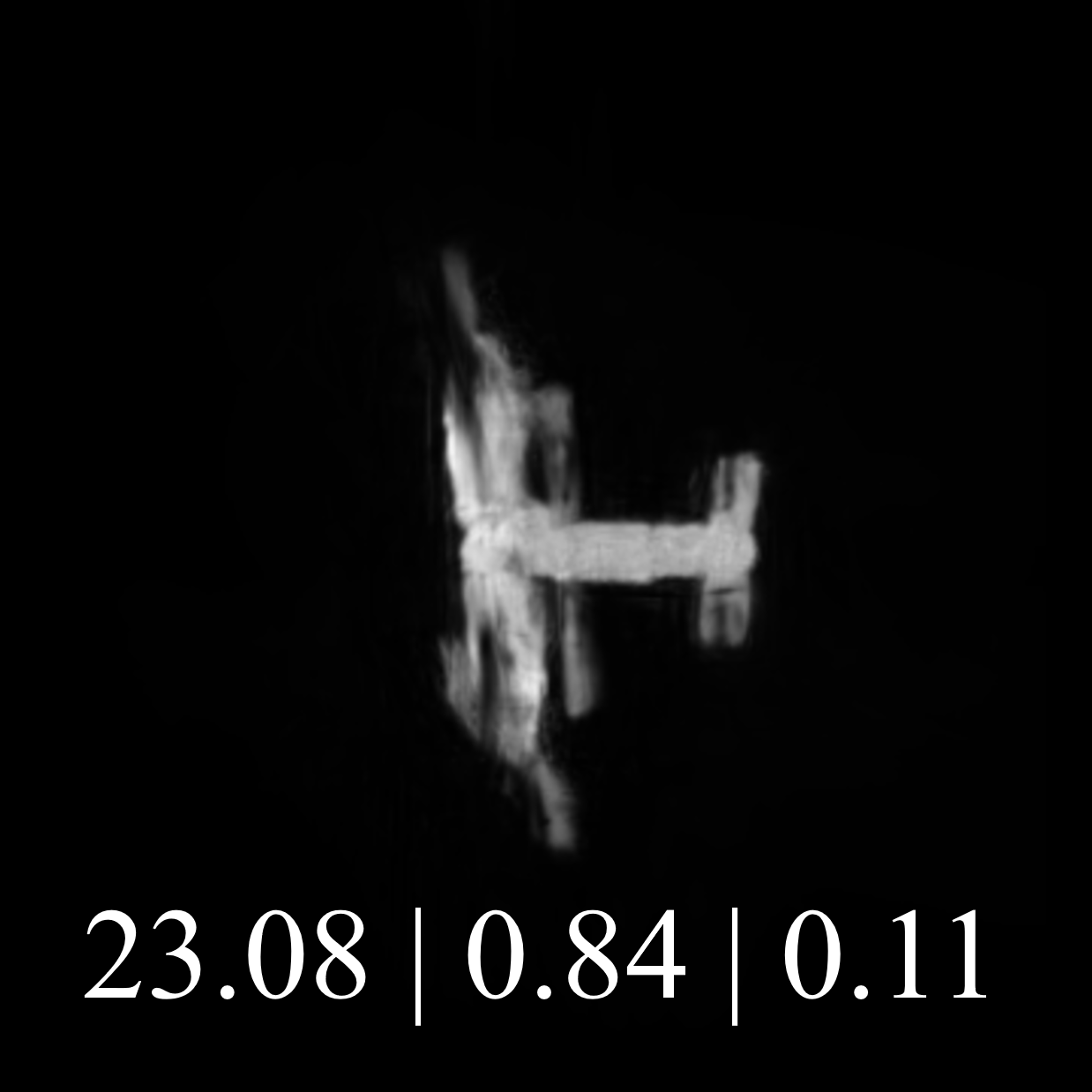}
            \subcaption{No Regulation}
        \end{minipage}%
        \begin{minipage}{0.14\textwidth}
            \includegraphics[width=\linewidth]{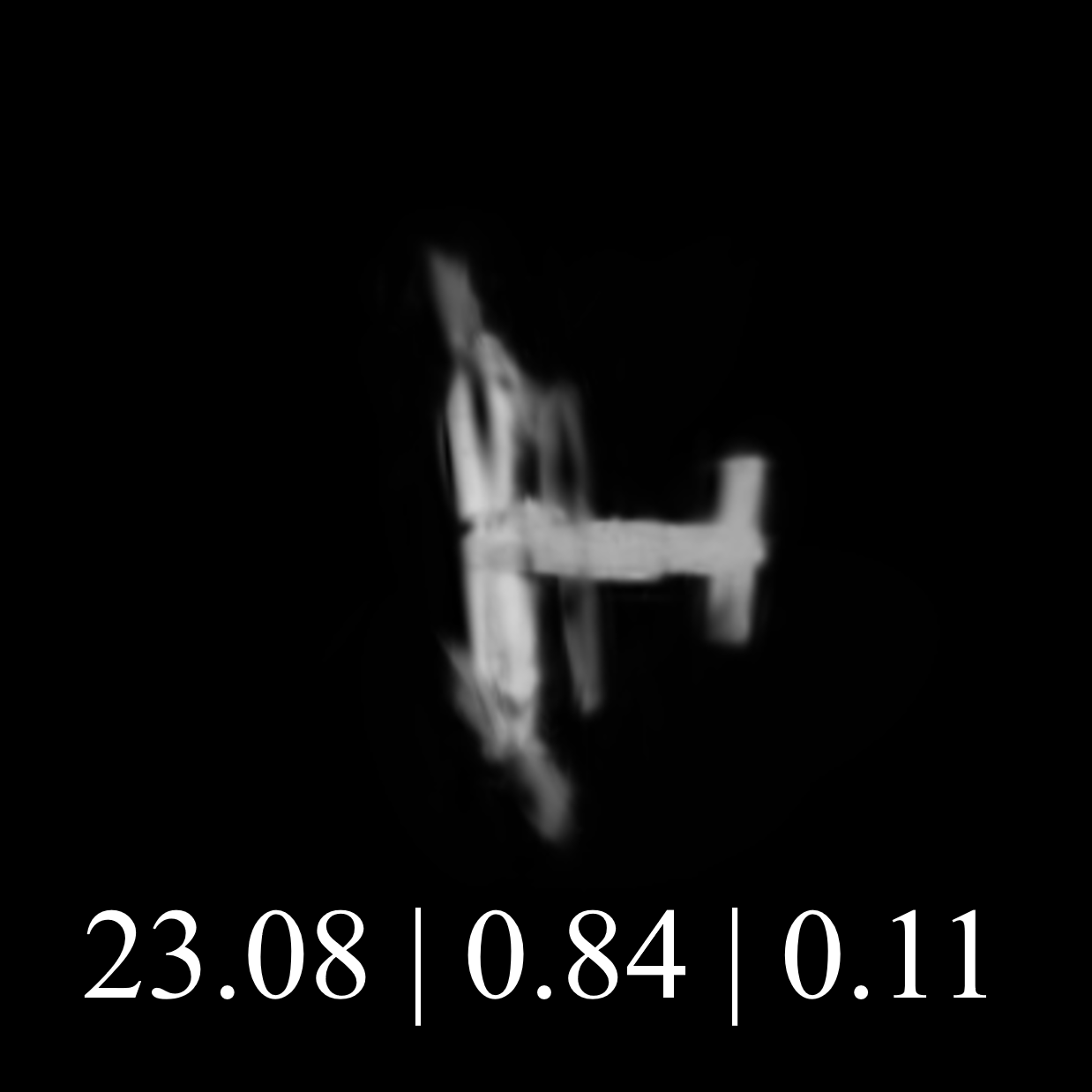}
            \subcaption{No BnB}
        \end{minipage}%
        \begin{minipage}{0.14\textwidth}
            \includegraphics[width=\linewidth]{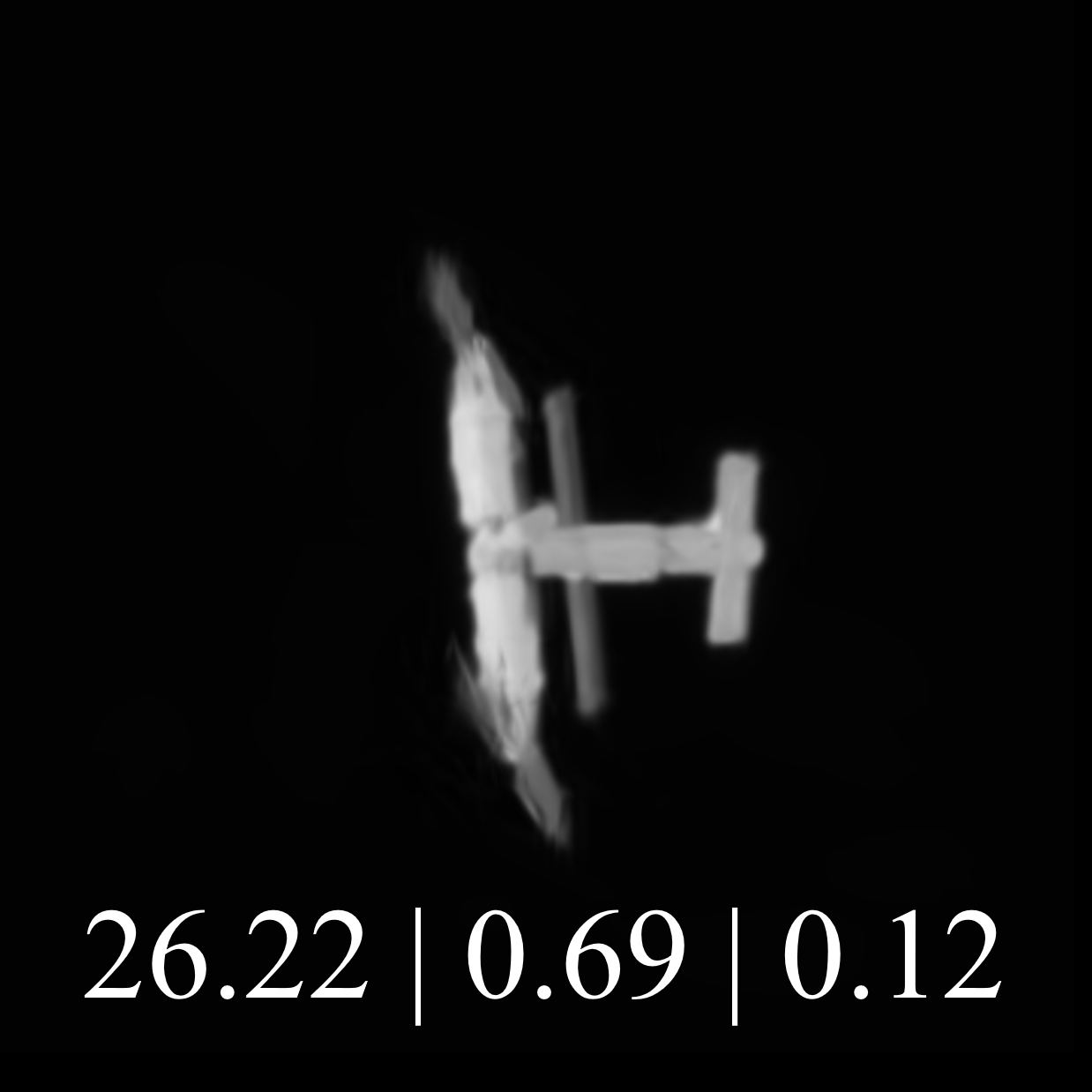}
            \subcaption{No Filtering}
        \end{minipage}
    \end{minipage}

    \caption{\textbf{Ablation studies of our 3D satellite reconstruction pipeline.} Reconstructed point clouds  (a) and novel views (b) of Simulation 2 data illustrate the effect of removing individual steps. (1) \textbf{Ours:} The complete pipeline produces accurate point cloud and novel views. (2) \textbf{Original GS:} Using the original GS algorithm—initialized with our modified SfM but lacking all subsequent steps—leads to overfitting and inaccurate pose estimation, resulting in chaotic point clouds and severe artifacts in novel views. (3) \textbf{No Preprocessing:} Omitting preprocessing yields noisy point clouds and overly-smoothed novel view renderings. (4) \textbf{No Regulation:} The absence of constraints on point cloud growth rate exacerbates overfitting, producing super noisy point clouds. (5) \textbf{No BnB:} Excluding BnB refinement introduces artifacts in both point clouds and novel view synthesis due to erroneous pose estimation. (6) \textbf{No Filtering:} Omitting point cloud filtering results in persistent background clutter and residual outliers. These studies underscore the critical role of each step in achieving stable, detailed 3D reconstructions and high-quality novel view synthesis.
    }
    \label{fig:ablation}
    
\end{figure*}

\setlength{\tabcolsep}{6pt}
\begin{table*}[!h]
  \centering
  \caption{\textbf{Ablation studies of our 3D satellite reconstruction pipeline on three simulation datasets.} Quantitative metrics (PSNR, SSIM, LPIPS, CD) for novel view synthesis are reported. Best results in each column are highlighted in \textbf{bold}, and second-best are \underline{underlined}.
  % Ablation study on three simulated data. For clarity, the results in the table represent novel view synthesis outcomes, with the optimal result in each column highlighted in \textbf{bold} and the second-best result \underline{underlined}.
  }
  \label{tab:ablation}
  \begin{tabular}{lcccccccccccc}
    \toprule
    &\multicolumn{4}{c}{Simulation 1} & \multicolumn{4}{c}{Simulation 2} & \multicolumn{4}{c}{Simulation 3}\\
    Method & \scriptsize{PSNR\textsuperscript{$\uparrow$}} & \scriptsize{SSIM\textsuperscript{$\uparrow$}} & \scriptsize{LPIPS\textsuperscript{$\downarrow$}} &
    \scriptsize{CD\textsuperscript{$\downarrow$}} &
    \scriptsize{PSNR\textsuperscript{$\uparrow$}} & \scriptsize{SSIM\textsuperscript{$\uparrow$}} & \scriptsize{LPIPS\textsuperscript{$\downarrow$}} & 
    \scriptsize{CD\textsuperscript{$\downarrow$}} &
    \scriptsize{PSNR\textsuperscript{$\uparrow$}} & \scriptsize{SSIM\textsuperscript{$\uparrow$}} & \scriptsize{LPIPS\textsuperscript{$\downarrow$}} &
    \scriptsize{CD\textsuperscript{$\downarrow$}}\\
    \midrule
    Ours & \textbf{25.33} & \textbf{0.90} & \underline{0.16} & \textbf{0.07} & \textbf{27.53} & \textbf{0.91} & \underline{0.14} & \textbf{0.08} & \textbf{28.10} & \textbf{0.91} & \underline{0.11} & \textbf{0.07}\\
    Original GS &  20.96 & 0.35 & 0.18 & 0.45 & 23.81 & 0.36 & 0.17 & 0.15 & 22.12 & 0.34 & 0.20 & 0.17\\
    No Preprocessing & 20.87 & \underline{0.87} & 0.22 & 0.38 & 25.76 & 0.37 & 0.21 &0.15 & 22.24 & \underline{0.85} & 0.22 &0.13 \\
    
    % No Sharpening & \underline{25.04} & \underline{0.89} & 0.19 & 0.28 & 32.42 & 0.53 & 0.11 & \textbf{0.02} & \textbf{36.64} & 0.54 & \textbf{0.01}& 0.12\\
    
    % No Gaussian initialization & 18.10 & \underline{0.89} & 0.25 &0.35 & 18.63 & 0.57 & 0.22 & 0.43 & 18.85 & 0.57 & 0.25 & 0.38\\
    No Regulation & 22.98 & \underline{0.87} & \textbf{0.14} &\underline{0.21} & 26.23 & 0.87 & \textbf{0.13} & 0.16 & 27.72 & 0.65 & \textbf{0.10} & 0.12 \\
    
    No BnB & 22.51 & 0.86 & 0.18 &0.30 & 25.89 & \underline{0.90} & 0.16 & \underline{0.14} & \underline{28.01} & \textbf{0.91} & 0.12 & \underline{0.09}  \\

    No Filtering & \underline{24.98} & 0.62 & 0.17 & 0.31 & \underline{26.89} & 0.61 & 0.16 & 0.16 & 27.88 & 0.63 & 0.12 & 0.14 \\
    \bottomrule
  \end{tabular}
\end{table*}

\subsubsection{Observations of CSS and ISS}
\label{subsubsec:Observationscssandiss}
The CSS was observed on September 15, 2023, from 20:48:30 to 20:51:17 UTC near Miyun Reservoir, Beijing, with a closest approach of 389.4 km and a peak elevation of 78.3°. The ISS was observed on August 21, 2024, from 20:20:06 to 20:22:09 UTC in Yanqing District, Beijing, with a closest approach of 442 km and a peak elevation of 86°. For both observations, we used a Celestron C14HD telescope (35cm aperture, f/11) with a QHY5III678M camera (shown in the lower right corner of Fig.~\ref{fig:css}), achieving a spatial sampling rate of 0.106"/pixel (0.2m/pixel at 400km). This sampling rate is smaller than the optical resolution: The diffraction limit of our 35 cm telescope is approximately 0.39", corresponding to 76cm resolution at 400km. Our observations employed the "Fake Polar Axis" method with a German equatorial mount (CGX-L), whose polar axis was tilted into the satellite's orbital plane rather than aligned with the celestial pole (see Fig.~\ref{fig:rebuttal} (b)). This orientation eliminates the need for meridian flips and enables uninterrupted tracking during the satellite pass. As most of the satellite's apparent motion is projected onto the declination (Dec) axis, only minor compensation is required from the right ascension (RA) axis, reducing field rotation—especially near the center of the image sequence. We implemented this method using a custom tracking program. Images were captured at 5 ms exposures, averaging 88 fps for CSS and 55 fps for ISS, resulting in datasets of 13,800 images for CSS and 6,770 for ISS, respectively. The relative angular sweeps between the telescope and the satellites span approximately 99° for CSS and 82° for ISS.

\subsubsection{3D Reconstruction and Metrology} 
We first apply the pre-processing pipeline to extract clean frames from the raw observations, selecting 138 clean frames for CSS and 116 for ISS. Following our simulation protocol, we chose one image out of every ten for training—yielding 14 images for CSS and 12 for ISS—with the remainder reserved for validation. Our method is again benchmarked against three NeRF-based baselines, NeRF$--$, BARF and BARF's variant (BARF with Constrained Sampling, abbreviated as BARF (Con.)).

The 3D reconstruction results for CSS and ISS (Fig.~\ref{fig:compare_real}) closely mirror our simulation findings. while NeRF-based methods struggle to generalize to unseen viewpoints with limited training data, our approach achieves more accurate 3D reconstruction and generated more realistic new views. Quantitative metrics reported in Table~\ref{tab:result} also demonstrates our method's superior performance.

We further validate our reconstructions via detailed metrology. Specifically, we measure the lengths and diameters of four CSS modules (Tianhe, Mengtian, Wentian, and Tianzhou-6) and determine the angles and dimensions of their solar arrays (see Fig.~\ref{fig:css}). Using our reconstructed 3D Gaussian splatting model, we densely sample 2D views to identify frames where each module appeared in orthogonal perspective at its maximum extent, then compute dimensions via pixel counts. The solar array angles are derived from their spatial relationships with the main modules. As summarized in Table~\ref{tab:measure}, our metrology results align with the officially released CSS specifications.

\subsection{Ablation Study}
We finally conduct comprehensive ablation studies on our simulation data to evaluate the contribution of each component in our 3D reconstruction framework. Fig.~\ref{fig:ablation} display reconstructed point clouds and representative novel view synthesis results for the naive Gaussian Splatting method and when individual steps are omitted. 

Each ablation visibly degrades reconstruction performance, highlighting the importance of every step in our pipeline for accurate pose estimation, high-fidelity 3D reconstruction, and consistent novel view synthesis. The absence of any component results in errors like blurriness, ghosting, structural discontinuities, fringe artifacts, or noisy backgrounds, underscoring the necessity of our complete method.

Quantitative evaluation using PSNR, SSIM, LPIPS, and CD metrics (see Table~\ref{tab:ablation}) confirms that every step in our pipeline contributes to reconstruction quality. Notably, the modified SfM-based initialization, controlled Gaussian growth, and BnB-search-based pose refinement are the most critical innovations in our method.

\section{Conclusion}
In this paper, we present a novel 3D imaging framework for reconstructing satellites from ground-based telescope observations. Our approach integrates a hybrid telescope image pre-processing pipeline with a joint pose-refinement and 3D reconstruction framework that leverages controlled Gaussian Splatting (GS) and Branch-and-Bound (BnB) search. We validated our method on synthetic datasets and on-sky observations, achieving, for the first time, 3D imaging of China’s Tiangong Space Station and the International Space Station using ground-based telescopes.

Due to hardware limitations, such as the telescope’s aperture size, our current 3D imaging resolution remains restricted. Additionally, our current model, based on a simple photometric-consistency loss, does not fully capture strong specular reflections from metallic surfaces and solar panels, which may bias the fit under highly specular conditions. In future work, we plan to use larger, more stable telescopes to improve image resolution and signal-to-noise ratio, enabling the reconstruction of smaller satellites. We will also develop modular interfaces so that updated denoising and matching networks can be seamlessly integrated. Additionally, we plan to incorporate material properties into our 3D satellite models to better simulate sunlight reflection and occlusion effects. We will further extend our approach to capture specular reflections by using reflectance-aware Gaussian splatting. Additionally, we will explore other applications of our method for 3D imaging at long observation distances and with low-quality images.

\vspace{-7pt}

% Any acknowledgments to only be included in camera ready

\section*{Acknowledgments}
This work was supported by National Natural Science Foundation of China (Grant No. 62371007, 62136001, 62088102, 12173041).

% \vspace{2cm}
\bibliographystyle{IEEEtran}
\bibliography{references}

\begin{IEEEbiographynophoto}{Zhiming Chang}
received the M.S. degree from the College of Engineering, Peking University, Beijing, China, in 2025. His research interests include computational imaging, 3D reconstruction, and autonomous driving.
\end{IEEEbiographynophoto}\vspace{-5cm}
\begin{IEEEbiographynophoto}{Boyang Liu}
received his Ph.D. degree from the International Centre for Radio Astronomy Research (ICRAR) at the University of Western Australia in 2021, and his bachelor's degree from Peking University in 2012. He is currently involved in the large optical telescope project EAST, initiated by Peking University. His interests also include public outreach in astronomy and the observational study of artificial objects in low Earth orbit.
\end{IEEEbiographynophoto}\vspace{-5cm}
\begin{IEEEbiographynophoto}{Yifei Xia}
is an M.S. student in the School of Computer Science at Peking University. He received the B.S. from Peking University in 2023. His research interests include computational photography and video generation.
\end{IEEEbiographynophoto}\vspace{-5cm}
\begin{IEEEbiographynophoto}{Youming Guo}
received the B.S. degree in automation from the University of Science and Technology of China, Anhui, China, in 2009 and the Ph.D. degree in signal and information processing from the Institute of Optics and Electronics, Chinese Academy of Sciences, Sichuan, China, in 2014. In 2014, he joined the Institute of Optics and Electronics, Chinese Academy of Sciences, where he is currently a Researcher and a Ph.D. Supervisor. He has authored or coauthored more than 70 papers in high-impact domestic and international journals. He has been granted more than 20 patents and was the principal investigator for projects including the National Natural Science Foundation of China (NSFC), high-tech initiatives, and innovation special zones. His primary research areas include adaptive optics, intelligent control, and computational imaging. Dr. Guo is a member of the Youth Innovation Promotion Association of the Chinese Academy of Sciences (CAS), a Western Young Scholar of CAS, and a council member of the Sichuan Youth Science and Technology Association.
\end{IEEEbiographynophoto}\vspace{-5cm}
\begin{IEEEbiographynophoto}{Boxin Shi}
received the BE degree from the Beijing University of Posts and Telecommunications, the ME degree from Peking University, and the PhD degree from the University of Tokyo, in 2007, 2010, and 2013. He is currently a Boya Young Fellow Associate Professor(with tenure) and Research Professor at Peking University, where he leads the Camera Intelligence Lab. Before joining PKU, he did research with MIT Media Lab, Singapore University of Technology and Design, Nanyang Technological University, National Institute of Advanced Industrial Science and Technology, from 2013to 2017. His papers were awarded as Best Paper,Runners-Up at CVPR 2024, ICCP 2015, and selected as Best Paper candidate at ICCV 2015. He is an associate editor of TPAMI/IJCV and an area chair of CVPR/ICCV/ECCV. He is a senior member of IEEE.
\end{IEEEbiographynophoto}
\vspace{-5cm}
\begin{IEEEbiographynophoto}{He Sun}
is an Assistant Professor at the National Biomedical Imaging Center, Peking University, China. 
Prior to joining Peking University, he was a Postdoctoral Researcher in the Department of Computing 
and Mathematical Sciences at the California Institute of Technology. He received the Ph.D. degree in 
Mechanical and Aerospace Engineering from Princeton University in 2019 and the bachelor's degree in 
Engineering Mechanics and Economics from Peking University in 2014. His research primarily focuses on computational imaging, which tightly integrates optics, control, signal processing and machine learning to push the boundary of scientific imaging. His past work has contributed to multiple real science missions, including the Event Horizon Telescope for black hole interferometric imaging and space telescope missions for exoplanet detection.
\end{IEEEbiographynophoto}

% \ifpeerreview \else


% \end{IEEEbiography}

%\newpage


%\vfill

% \fi

\end{document}